\documentclass{article}

\PassOptionsToPackage{numbers,compress}{natbib}
\usepackage[main,final]{neurips_2026}
\usepackage{longtable}
\usepackage{xcolor}
\definecolor{cgreen}{rgb}{0.2,0.6,0.2}
\definecolor{darkred}{rgb}{0.4,0.0,0.0}
\definecolor{darkgreen}{rgb}{0.0,0.4,0.0}
\definecolor{darkblue}{rgb}{0.0,0.0,0.4}
\definecolor{inexposure}{HTML}{4B9DA9}
\definecolor{outofscope}{HTML}{E37434}

\usepackage[colorlinks=true,  linkcolor=darkred, urlcolor=darkblue, citecolor=darkblue]{hyperref}  

\usepackage[utf8]{inputenc} % allow utf-8 input
\usepackage[T1]{fontenc}    % use 8-bit T1 fonts
\usepackage{hyperref}       % hyperlinks
\usepackage{url}            % simple URL typesetting
\usepackage{booktabs}       % professional-quality tables
\usepackage{amsfonts}       % blackboard math symbols
\usepackage{nicefrac}       % compact symbols for 1/2, etc.
\usepackage{microtype}      % microtypography
\usepackage{xcolor}         % colors
\usepackage[table]{xcolor}
\usepackage{soul}
\usepackage{graphicx}
\usepackage{svg}
\usepackage{changepage}
\usepackage{enumitem}
\usepackage{amsmath} 
\usepackage{cleveref}
\usepackage{etoc}
\usepackage{todonotes}
\usepackage{fontawesome5}

\title{LittleLearner: Language Models Under Pedagogically Controlled Knowledge Exposure}

\definecolor{llblue}{HTML}{68A8F1}
\definecolor{llblueLight}{HTML}{7DC9FF}
\definecolor{llblueDark}{HTML}{2F4A6D}

\definecolor{llcoral}{HTML}{FD7976}
\definecolor{llcoralLight}{HTML}{FFCC8F}
\definecolor{unfiltered}{HTML}{fcb18e}
\definecolor{littlelearner}{HTML}{79b2f2}
\definecolor{llcoralDark}{HTML}{7A2E2C}

\usepackage[breakable]{tcolorbox}
\newcommand{\callout}[1]{%
    \begin{center}
    \begin{tcolorbox}[colframe=llblueLight, colback=llblueLight!30, coltext=llblueLight!40!black, width=\columnwidth, left=.7mm, right=.7mm, top=1mm, bottom=1mm, boxrule=.5mm]
        #1
    \end{tcolorbox}
    \end{center}
}

\newcounter{takeaway}
\newcommand{\takeaway}[1]{%
    \callout{#1}
}

\newcommand{\prompt}[2]{%
\begin{tcolorbox}[breakable]
\textbf{#1}\\
{\ttfamily
\small
#2
}
\vspace{0.8em}
\end{tcolorbox}
}

\newcommand{\LittleLearner}[0]{\scalebox{0.9}[1.0]{\textsc{LittleLearner}}}
\newcommand{\General}[0]{\scalebox{0.9}[1.0]{\textsc{Unfiltered}}}
\newcommand{\LittleCurriculum}[0]{\scalebox{0.9}[1.0]{\textsc{LittleCurriculum}}}
\newcommand{\KFive}[0]{\textcolor{inexposure}{\textbf{K--5}}}
\newcommand{\BKFive}[0]{\textcolor{outofscope}{\textbf{Beyond-K--5}}}
\newcommand{\FinewebEdu}[0]{FineWeb-Edu}
\newcommand{\CommonCoreText}[0]{CommonCoreText}

\author{
Fanfei Li\thanks{Equal contribution. Correspondence to
\texttt{fanfei.li@tuebingen.mpg.de}.} \\
MPI-IS, Ellis Institute
\And
Jana Zeller\footnotemark[1]\\
MPI-IS, Ellis Institute, ETHZ
\And
Manuel Prada-Corral \\
MPI-IS, Ellis Institute, ETHZ
\And
Thaddäus Wiedemer \\
MPI-IS, Ellis Institute
\And
Prasanna Mayilvahanan \\
MPI-IS, Ellis Institute
\And
Ryan Cotterell \\
ETHZ
\And
Wieland Brendel \\
MPI-IS, Ellis Institute
}

\begin{document}

\maketitle
% \vspace{-2.0 em}
% \begin{center} \href{https://littlelearner-ll.github.io/}{\faGlobe\ \textbf{LittleLearner}} \end{center} 

\begin{abstract}
Modern language models are trained on heterogeneous web-scale text corpora.
Consequently, studying knowledge and skill acquisition is difficult, as prior exposure to related content is hard to characterize.
To address this challenge, we introduce \textbf{\LittleCurriculum{}}, a curated 88B-token pretraining corpus tailored to U.S. elementary school material, explicitly excluding concepts, facts, and vocabulary taught above Grade 5.
Training a 5B-parameter large language model from scratch on \LittleCurriculum{} yields \textbf{\LittleLearner{}}, a model with sufficient language competence for open-ended evaluation, yet with clear knowledge and capability boundaries mapped to interpretable curriculum guidelines.
We release \LittleCurriculum{} and \LittleLearner{} as a developmentally restricted sandbox to study how models acquire, represent, and use data under a well-defined training scope.
We illustrate the sandbox's utility in a first suite of experiments on injecting new knowledge through post-training and in-context learning. These methods let \LittleLearner{} better utilize existing knowledge, but do not raise out-of-scope capabilities.
Our findings underscore the value of this controlled environment for future investigations.
Models and further details are available on the \href{https://littlelearner-ll.github.io}{project page}.
\end{abstract}

\begin{figure}[h]
    \centering
    \includegraphics[width=\linewidth]{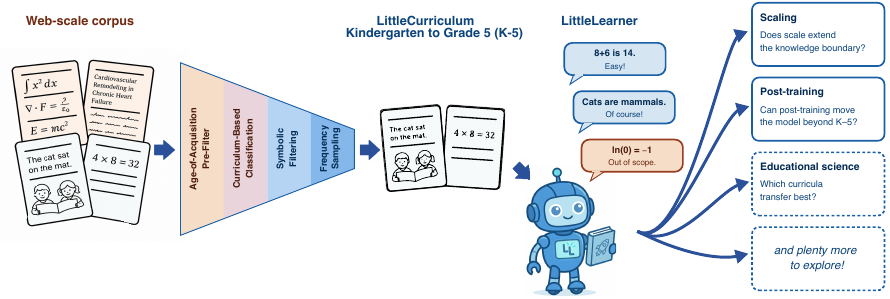}
    \caption{\textbf{\LittleCurriculum{} and \LittleLearner{} provide a pedagogically grounded sandbox for studying model behavior under precise data constraints}.
    \LittleCurriculum{} is an 88B-token pretraining corpus filtered from \FinewebEdu~\citep{lozhkov2024fineweb-edu} to only contain elementary school (\KFive{}) data. On it, we train a 5B model: \LittleLearner{}. Together, they form a well-curated sandbox to study scaling, post-training, educational science and more.}
    \label{fig:teaser-figure}
\end{figure}

\section{Introduction}
\label{sec:intro}
Large language models (LLMs) acquire capabilities from massive web-scale text corpora spanning a myriad of heterogeneous sources, making their prior knowledge difficult to characterize. A growing body of evidence suggests that model behavior is strongly shaped by this pretraining exposure: data contamination can inflate benchmark performance~\citep{yang2023rethinking,deng2024investigating,dominguezolmedo2025trainingtesttaskconfounds,mayilvahanan2025searchforgottendomaingeneralization,mayilvahanan2025llmslinedatadetermines}, phenomena such as the reversal curse reveal brittle generalization from observed facts~\citep{berglund2024reversalcurse}, and targeted data patches are often needed to repair missing skills~\citep[e.g.,][]{xia2024less,chen2023skillit,lee2024llm2llm}. When prior exposure is unknown, it is difficult to determine whether methods like in-context learning or post-training lead to genuine capability growth or merely elicit knowledge or skills already present in the training data~\citep[cf.][]{zhou2023lima,workrethinking,pan2023context,wang2026beyond,vergara2026operationalising}.

Prior work has largely addressed this problem through benchmark design, using increasingly difficult and carefully curated out-of-distribution evaluations~\citep[e.g.,][]{wu2024reasoning,mayilvahanan2025}. While valuable, such evaluations provide only indirect control over prior exposure: novelty must be reassessed for each new evaluation, and knowledge boundaries remain difficult to conceptualize and verify. We take a complementary approach: instead of defining novelty at evaluation time, we constrain the training distribution itself.

Although the cut-off date for pretraining corpora serves as an implicit \textit{temporal} knowledge boundary, which \citet{levine2026talkie} study concurrently in more detail, this only limits what information is available, not the specialized vocabulary and (historical, albeit sophisticated) scientific or mathematical concepts the model sees.
Instead, we focus on \textit{developmental} boundaries that more directly constrain exposure, as age-level curricula provide a pedagogically grounded proxy for the concepts, language, and reasoning demands available at a given developmental stage.

We instantiate this approach with \LittleLearner{}, a 5B-parameter language model trained from scratch under controlled knowledge exposure. \LittleLearner{} is trained on \LittleCurriculum{}, a developmentally constrained 88B-token corpus corresponding to the U.S. kindergarten to elementary school curriculum (denoted as \KFive{}), which we construct using a scalable filtering pipeline.
The model and dataset form a controlled sandbox for studying how restricted pretraining exposure shapes learning and later capability growth.
% As illustrated in \Cref{tab:qualitative-base}, the model produces fluent conversational completions, while out-of-scope prompts expose conceptually misplaced or confabulatory behavior.
In particular, this allows us to study whether in-context learning~(ICL) and post-training \emph{extend} a model's capabilities beyond pretraining or merely \emph{elicit} latent knowledge.
 More broadly, \LittleLearner{} supports principled studies of how language models acquire, transfer, calibrate, and extend knowledge under explicitly constrained pretraining exposure.

\begin{table}[h]
\centering
\setlength{\tabcolsep}{6pt}
\renewcommand{\arraystretch}{1.08}
\caption{\textbf{\LittleLearner{} responds appropriately for an elementary school student compared to an \General{} baseline}.
Qualitative examples on in-scope \KFive{} and out-of-scope \BKFive{} questions. 
Refer to \Cref{app:qualitative-base} for more examples.}
\label{tab:qualitative-base}
\small
\begin{tabular}{p{0.48\textwidth} p{0.48\textwidth}}
\toprule
\textbf{\LittleLearner{}} (only exposure to \KFive{}) & \textbf{\General} (control with unfiltered exposure) \\
\midrule
\multicolumn{2}{l}{\textcolor{inexposure}{\textbf{In-scope} (\KFive{})\quad What is gravity?}}\\
Gravity is the force that pulls everything down.
&
Gravity is the force between any two objects that is inversely proportional to the square of the separation between them...
\\
\addlinespace[1ex]
\multicolumn{2}{l}{\textcolor{outofscope}{\textbf{Out-of-scope} (\BKFive{})\quad What is Schrödinger's cat?}}\\
Schrödinger's cat is a cat with two faces.
&
Schrödinger's cat is a thought experiment in quantum mechanics, proposed by Austrian physicist Erwin Schrödinger in 1935...
\\
\bottomrule
\end{tabular}
\end{table}

\takeaway{
In summary, we provide
\begin{enumerate}[nosep,before=\vspace{1ex},itemsep=1ex,rightmargin=0pt,leftmargin=2em]
\item \textbf{\LittleCurriculum{}}, a curated 88B-token dataset filtered to only contain elementary school (\KFive{}) content (\Cref{sec:data-filtering}).
\item \textbf{\LittleLearner{}}, a 5B model trained on \LittleCurriculum{}. The model and dataset form a sandbox for probing data acquisition, representation, and utilization within a well-defined training scope (\Cref{sec:model-training}).
\item First experiments with \LittleLearner{} showing that increased model size, post-training, and ICL do not improve out-of-scope performance far beyond training exposure, but can increase performance within scope and at the boundary (\Cref{sec:first-experiments-with-ll}).
\end{enumerate}
}

\section{Related Work}
\label{sec:rel}
% Our work connects to research that studies how pretraining data shapes language model behavior and to approaches that aim to disentangle genuine capability acquisition from the elicitation of knowledge already present in pretraining.

\paragraph{Knowledge Boundaries in LLMs}
Related work characterizes the knowledge boundary of an already-trained model~\cite{li2025knowledge}, probing what a fixed LLM does and does not know through self-knowledge and unanswerable-question detection, confidence calibration, retrieval-augmented behavior shifts, and prompt-sensitivity-based formalizations~\cite{kaur2026knowing, kadavath2022language, ren2025investigating, yin2024benchmarking}. Operating over an opaque pretraining distribution, these methods infer the boundary after training. Our setup instead specifies it in advance by controlling the training corpus, allowing mechanisms to be interpreted against known exposure.

\paragraph{Pretraining data dominates downstream behavior}
A growing body of work shows that the pretraining distribution is the primary determinant of generalization and downstream capability, with architecture and optimization playing comparatively minor roles~\citep{mayilvahanan2025searchforgottendomaingeneralization,mayilvahanan2025llmslinedatadetermines,hardt2026emerging,fang2022datadeterminesdistributionalrobustness}. Recent analyses of reinforcement learning further suggest that observed gains often elicit reasoning patterns already present in pretraining rather than introduce genuinely new capabilities~\citep{yue2025doesreinforcementlearningreally,zhao2025echochamberrlposttraining,zhang2025interplay}. 
In the same vein, \citet{dominguezolmedo2025trainingtesttaskconfounds} show that training on the test task inflates benchmark scores and confounds emergent-capability claims, as exposure to the evaluation distribution becomes indistinguishable from genuine expansion.
These findings motivate our approach to explicitly constrain the pretraining distribution to cleanly separate capability acquisition from latent prior knowledge.

\paragraph{Evaluation beyond training distribution}
A complementary line of work targets the same confounder from the evaluation side, curating benchmarks designed to minimize overlap with web-scale pretraining, such as Humanity's~Last~Exam~\citep{hle2026}, GPQA~\citep{rein2023gpqa}, MMLU-Pro~\citep{wang2024mmlu}, and MATH-B~\citep{mayilvahanan2025}. As mentioned in \Cref{sec:intro}, these provide only indirect control, the underlying training distribution remains opaque, and benchmarks must be continuously refreshed to stay informative.

\paragraph{Constrained pretraining distributions}
The closest line of work directly constrains the pretraining corpus. BabyLM~\citep{warstadt2023babylm}
% \ryan{It might make sense to cite the entire line of papers and discuss how each one added improvements over the previous one.}
restricts training to human-scale corpora to study sample efficiency and linguistic generalization, but restricts \emph{quantity} rather than \emph{conceptual scope}, leaving the model free to encounter advanced material. Concurrent work~\citep{levine2026talkie} imposes a \emph{temporal} cut-off by training on pre-1931 English text, yielding a clean historical boundary but no constraint on conceptual complexity within that period. \LittleCurriculum{} complements both by imposing a \emph{developmental} cut-off: the corpus is filtered to U.S. \KFive{} material along pedagogically grounded curriculum standards, giving an interpretable conceptual boundary at LLM scale.

\section{Constructing the \LittleLearner{}-\LittleCurriculum{} Sandbox}

\subsection{Constructing a Developmentally Restricted \LittleCurriculum}
\label{sec:data-filtering}
We divide the web corpus \FinewebEdu{}~\citep{lozhkov2024fineweb-edu} into an elementary school (\KFive{}) and \BKFive{} subset, via multi-stage filtering that enforces constraints on knowledge, syntactic complexity, and reasoning.

\subsubsection{Age-of-Acquisition Pre-filtering}
We begin with a lightweight rule-based filter to enforce coarse age alignment.
Among several proxies for linguistic and conceptual difficulty, Age-of-Acquisition~(AoA)~\citep{kuperman2012age} provides the strongest separation between grade levels; see \Cref{app:rule-based-pre-filtering}.
However, 10\% of the words in \FinewebEdu{} are not covered by AoA.
Therefore, we impute missing AoA values using a linear model over the log word frequency of Zipf WordFreq~\citep{robyn_speer_2022_7199437}, leveraging the reported relationship between AoA and frequency by~\citet{kuperman2012age}.
To be more robust to single-word outliers, we only discard samples in which more than 5\% of the words exceed the target age of 12.

\subsubsection{LLMJ Annotation and Classifier Training}
We employ LLM-as-a-judge (LLMJ) to annotate a subset of \FinewebEdu{}, which serves as training data for downstream classifiers.
We initialize prompts based on Common Core State Standards~(CCSS)~\citep{ccss2010} and refine them through automatic prompt optimization using DSPy~\citep{khattab2024dspy} and OpenEvolve~\citep{openevolve}; see \Cref{app:llmj_setup}.
Note that full LLMJ annotation is prohibitively expensive: based on Gemini Flash pricing, labeling the full \FinewebEdu{} dataset would cost approximately USD~46M, further motivating our multi-stage approach.

\subsubsection{Curriculum-Based Classification}
Our filtering pipeline is centered on classifiers trained on LLMJ-annotated data.
We first apply a lightweight FastText~\citep{joulin2017bag} classifier. For the \textasciitilde266M samples whose grade band matches the AoA assignment, we apply a stronger, 50× more expensive ModernBERT classifier~\citep{warner2024smarterbetterfasterlonger}.
% We retain only samples for which both classifiers agree on the grade label.

\begin{figure}[t]
    \centering
    \includegraphics[width=\linewidth]{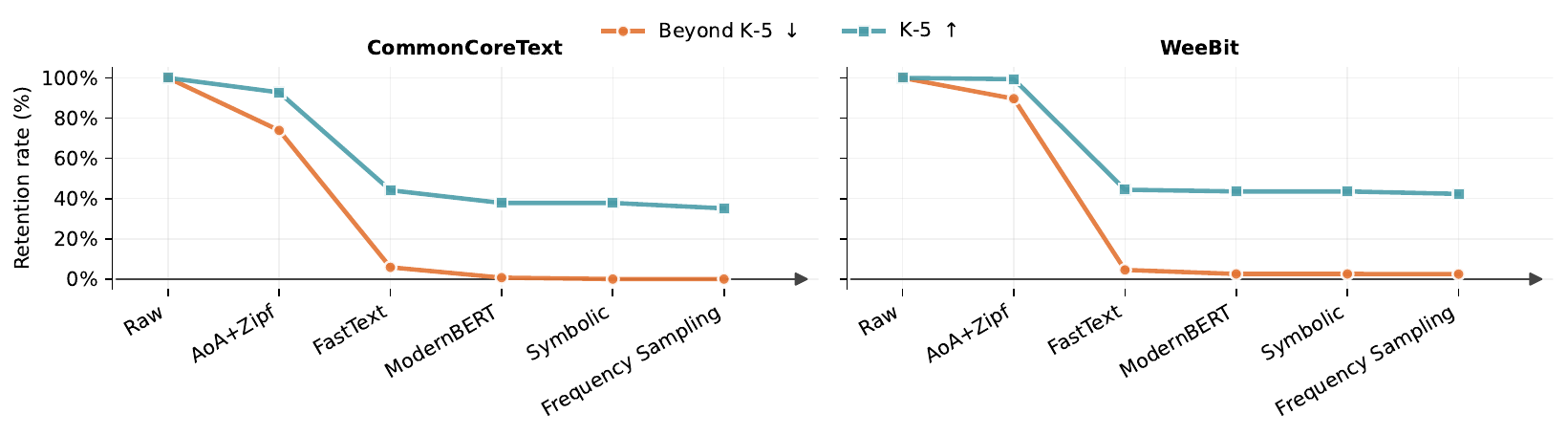}
    \caption{\textbf{The filtering yielding \LittleCurriculum{} conservatively enforces the \KFive{} boundary}.  Per-stage retention of \KFive{} and \BKFive{} documents on \CommonCoreText{} (left) and WeeBit (right). CommonCoreText contains CCSS-aligned reading materials (see \Cref{app:ccss-construction}) used during the construction of the filtering pipeline. WeeBit is an external benchmark with grade-level labels. On both benchmarks the pipeline achieves a \BKFive{} document retention of near-zero while retaining 35--42\% of \KFive{} documents, reflecting a deliberate \emph{precision-first} design: we sacrifice recall to obtain a sharp, interpretable knowledge boundary.}
    \label{fig:filtering-funnel}
\end{figure}

\subsubsection{Symbolic Filtering}
Due to the low frequency of formulas and the difficulty in learning reliable signals from LLMJ annotations, we apply a rule-based symbolic filtering step. Each document is scanned using a fixed set of regular expressions covering broad notation families, such as quadratic expressions and operator symbols (e.g., $\sum$, $\Sigma$, $\int$, $\partial$). To prioritize precision, we conservatively filter a document upon any match, as recognizable algebraic expressions are typically \BKFive. This step removes an additional \textasciitilde0.1\% of documents retained after the preceding classifier stage, suggesting that symbolic filtering primarily addresses a small residual tail of notation-heavy content rather than serving as a broad mathematics filter. See \Cref{app:symbolic-filtering} for details.

\subsubsection{Frequency Sampling}
The post-filter corpus is larger than our training target, allowing the final sampling stage to further tighten the exposure boundary rather than subsample uniformly. Using the grade assignments from the preceding stages, we identify terms that are disproportionately associated with \BKFive\ material and remove candidate documents containing these terms.
This conservative filtering step is further motivated by 
\citet{ruis2024procedural}, who show that models can acquire factual knowledge from sparse occurrences without degrading downstream reasoning.
See \Cref{app:frequency-sampling} for details.

\subsubsection{Validation}
We validate our pipeline using \CommonCoreText{}, a held-out ground-truth dataset constructed from publicly available recommended reading materials and textbooks paired with grade-level labels (see \Cref{app:ccss-benchmark}).
Qualitative filtered examples are provided in \Cref{app:exp_dropped_samples}.
% We validate our pipeline on \CommonCoreText{} retention rates.
The pipeline successfully reduces the retention of \BKFive{} content to 0\% while preserving \textasciitilde35\% of \KFive{} (\Cref{fig:filtering-funnel}, left). This low \KFive{} retention reflects our deliberate precision-over-recall design: \LittleCurriculum{} is intended as a sharply bounded corpus for controlled exposure experiments rather than an exhaustive sample of elementary school material.
Importantly, this 65\% rejection rate is measured on ground-truth \KFive{} passages in the CommonCoreText benchmark. We adopt this precision-first design because recovering more \KFive{} material from the much noisier FineWeb-Edu corpus would require relaxing the filter, thereby reducing confidence that the retained corpus respects the \BKFive{} boundary.
% Final retention is consistent across science and literature, indicating that there is no domain-specific bias. 
This interpretation is supported by a manual comparison of retained and removed \KFive{} documents: removed passages are systematically longer and lexically more complex than retained ones. 
Full statistics are provided in \Cref{app:ccss-k5-analysis}.

As an independent check, we apply the same pipeline to WeeBit \citep{vajjala-meurers}, an external grade-labeled corpus (\Cref{fig:filtering-funnel}, right). The filter retains 2.48\% of 6,000 \BKFive{} passages; manual inspection identifies genuinely out-of-scope concepts in only three passages, corresponding to 0.05\% of the \BKFive{} split (the remaining passages are web-scraping artifacts or content without educational value). We further conduct a corpus-wide scan using 126 \BKFive{} n-grams drawn from curriculum standards, finding matches in only 0.09\% of retained passages. See Section~\ref{app:leakage-audits} for details.

\takeaway{
\LittleCurriculum{} is a \KFive-targeted subset of \FinewebEdu, built via a precision-first multi-stage filtering pipeline, verified on a CCSS-aligned dataset (CommonCoreText) and corroborated on an external grade-labeled corpus (WeeBit). See \Cref{fig:filtering-funnel}.
}

\subsection{Training \LittleLearner{}}
\label{sec:model-training}
Using the constructed datasets, we train a 5B-parameter language model from scratch, which we call \LittleLearner{}.
We follow the Qwen3~\citep{yang2025qwen3technicalreport} architecture and pretrain the model for 100 hours on 8 NVIDIA B200 GPUs.
Refer to \Cref{app:model-training} for details on model training and parameter count.

\subsection{Validating \LittleLearner's Knowledge Exposure}
We validate the limited exposure of \LittleLearner{} via language complexity and performance on downstream benchmarks.
We compare against a model of the same size and type trained on an unfiltered version of \FinewebEdu{} (\General{}) and Gemma~2B~\citep{team2024gemma} as an external model of similar size, but trained on 2T instead of 80B tokens.

\subsubsection{Language Familiarity Validation}
% We assess \LittleLearner{}'s familiarity with \KFive{} and \BKFive{} language on two datasets.

\begin{figure*}[t]
    \centering
        \begin{minipage}{0.48\textwidth}
        \centering
        \includegraphics[width=\linewidth]{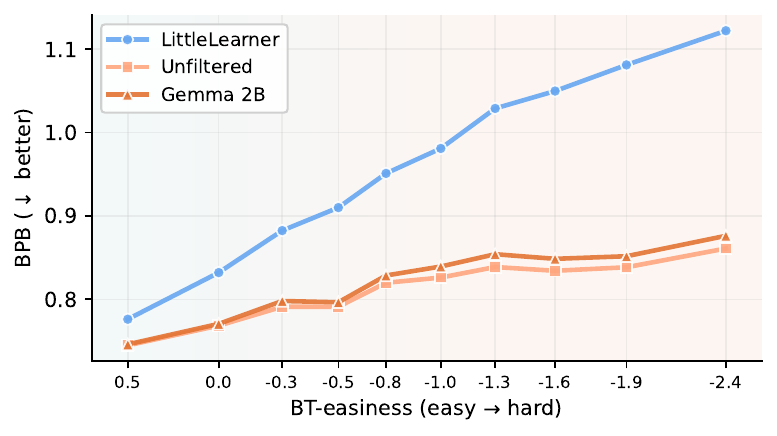}
        \caption{\textbf{\LittleLearner{} is increasingly unfamiliar with more difficult text}, as shown by increasing bits-per-byte~(BPB)~\citep{xue2022byt5} on CLEAR~\citep{crossley2023large}. Samples are binned by the pedagogical Bradley--Terry easiness score~\citep{bradley_terry}.}
        \label{fig:language-bpb}
    \end{minipage}
    \hfill
        \begin{minipage}{0.48\textwidth}
        \centering
        \includegraphics[width=\linewidth]{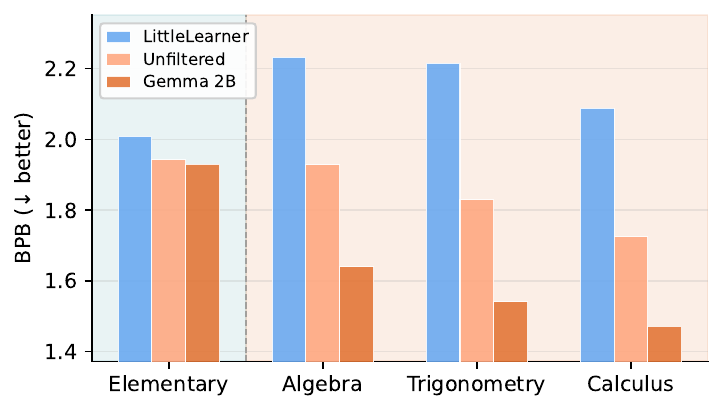}
        \caption{\textbf{\LittleLearner{} is unfamiliar with higher-grade mathematical content}, as indicated by BPB on student responses from student-assistant dialogues on mathematical discourse taken from CoMTA~\citep{khantutoring2024comta}.}
        \label{fig:math-bpb}
    \end{minipage}
\end{figure*}

\paragraph{Linguistic Complexity}
In \Cref{fig:language-bpb} we evaluate linguistic complexity using the CLEAR dataset~\citep{crossley2023large}, which provides Bradley–Terry~\citep{bradley_terry} easiness scores for text passages derived from expert teacher judgments.
\LittleLearner{}’s bits-per-byte (BPB)~\citep{xue2022byt5} steadily rises with increased hardness, while \General{}'s and Gemma~2B's BPB remain comparatively flat and closely track each other.
This divergence can be attributed to pretraining data, since \General{} shares \LittleLearner{}'s training recipe but not its training data.

\paragraph{Mathematical Complexity}
We validate a stronger mathematical familiarity on \KFive{} and a limited exposure on \BKFive{} on CoMTA~\citep{khantutoring2024comta} (\Cref{fig:math-bpb}).
This dataset captures real student responses in live tutoring sessions in different math categories, which we map to \KFive{} and \BKFive{} (see \Cref{app:comta-binning}).
We evaluate BPB only on the student turns, making sure to specifically test the mathematical familiarity of the models.
In \KFive{}, \LittleLearner{} reports a BPB similar to the two reference models.
However, in \BKFive{} \LittleLearner{} exhibits a growing unfamiliarity, while the unrestricted reference models remain stable.

\subsubsection{Capability Validation}
% We validate the expected capability cut-off induced by \KFive{} limited training in factual knowledge and mathematical reasoning abilities in \Cref{fig:mathcamps-pass-k-curve,fig:jeopardy-bar}.

\begin{figure*}[t]
    \centering
        \begin{minipage}{0.48\textwidth}
        \centering
        \includegraphics[width=\linewidth]{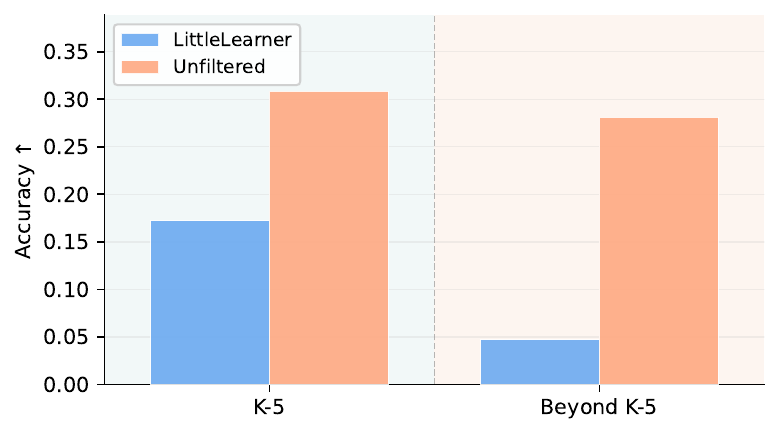}
        \caption{\textbf{\LittleLearner{} answers \KFive{} scientific questions but collapses on \BKFive}. Pass@1 accuracy on Jeopardy science questions~\citep{jeopardy_kaggle_tunguz} split by NGSS-aligned grade scope.}
        \label{fig:jeopardy-bar}
    \end{minipage}
    \hfill
        \begin{minipage}{0.48\textwidth}
        \centering
        \includegraphics[width=\linewidth]{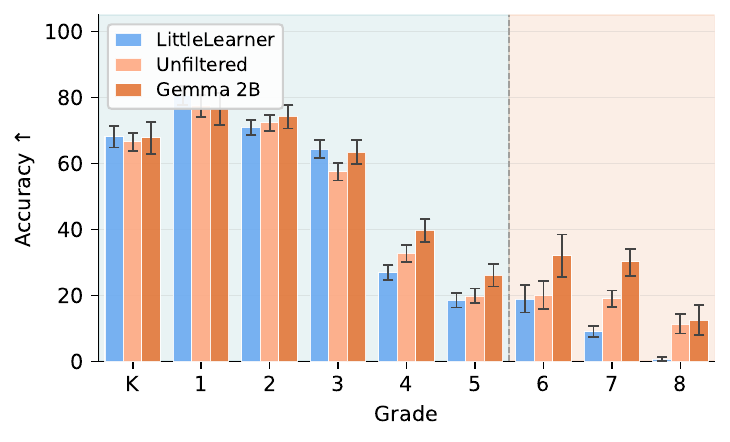}
        \caption{\textbf{\LittleLearner{} answers math questions on par in \KFive{}, but disproportionately collapses at higher difficulties}. Pass@1 accuracy on MathCAMPS estimated from 1024 rollouts. Whiskers indicate 95\% CIs.}
        \label{fig:mathcamps-pass-k-curve}
    \end{minipage}
\end{figure*}

\paragraph{Factual Knowledge}
We validate \LittleLearner's ability to retrieve factual knowledge within and beyond its training scope on Jeopardy science questions~\citep{jeopardy_kaggle_tunguz} divided into \KFive{} and \BKFive{} (see \Cref{fig:jeopardy-bar}). Gemini~Flash~\citep{Gemini2_5FlashModelCard} annotates each question with its corresponding NGSS (Next Generation Science Standards) curriculum strand~\citep{ngss2013}. We then split Jeopardy into questions within the NGSS \KFive{} curriculum and those at the \BKFive{} level. 
\LittleLearner{} shows a sudden drop in factual knowledge beyond its exposure horizon, while \General{} performs comparably across both bins. Since both models are trained using the same training recipe, this drop can be attributed to the limited knowledge available in \LittleCurriculum{}.

\paragraph{Mathematical Reasoning}
We characterize the mathematical reasoning boundary of \LittleLearner{} using MathCAMPS~\citep{mishra2024next}, a dataset of synthetic questions aligned with individual Common Core State Standards (CCSS) (\Cref{fig:mathcamps-pass-k-curve}). We adapt the dataset for model evaluation as described in \Cref{app:mathcamps-filtering}. 
Compared with \General{} and Gemma 2B, \LittleLearner{} exhibits a disproportionately large decline in performance at higher grade levels.
Increasing the sampling budget does not eliminate this gap: on Grade 8 problems, even at pass@1024, \LittleLearner{} solves fewer than half as many questions as \General{} (\Cref{app:mathcamps-pass-k}).
Rather than emerging at a single grade-level threshold, the performance gap gradually widens as the difficulty of the problem increases.
This trend partly reflects the limitations of grade-level CCSS labels as a proxy for problem difficulty for language models. Adjacent standards can target the same underlying operation while differing mainly in expected fluency, a distinction that is pedagogically meaningful, but may not translate meaningfully to language models. More generally, \LittleLearner{}'s capabilities do not follow the ordering of a human curriculum. For example, it performs better on division by a multi-digit divisor than on division by a single-digit divisor, despite single-digit division being a prerequisite to multi-digit division in human curricula (\Cref{app:mathcamps-learning-paths}).

\takeaway{
\LittleLearner’s abilities tightly track its controlled knowledge exposure: it performs strongly on in-scope language, math, and factual tasks, but degrades beyond this scope. See \Cref{fig:math-bpb,fig:language-bpb,fig:mathcamps-pass-k-curve,fig:jeopardy-bar}.
}

\section{Probing Capabilities with \LittleLearner{}}
\label{sec:first-experiments-with-ll}

\LittleLearner’s controlled pretraining enables a direct study of whether capabilities can be extended beyond a knowledge boundary or remain tied to pretraining exposure. We use the sandbox to examine three immediate adaptation questions that are difficult to isolate in standard web-scale models: whether increased scale expands the boundary, whether post-training can move the model beyond its \KFive{} scope, and whether in-context examples can recover out-of-scope \BKFive{} performance.

\begin{figure*}[t]
    \centering
    \includegraphics[width=\linewidth]{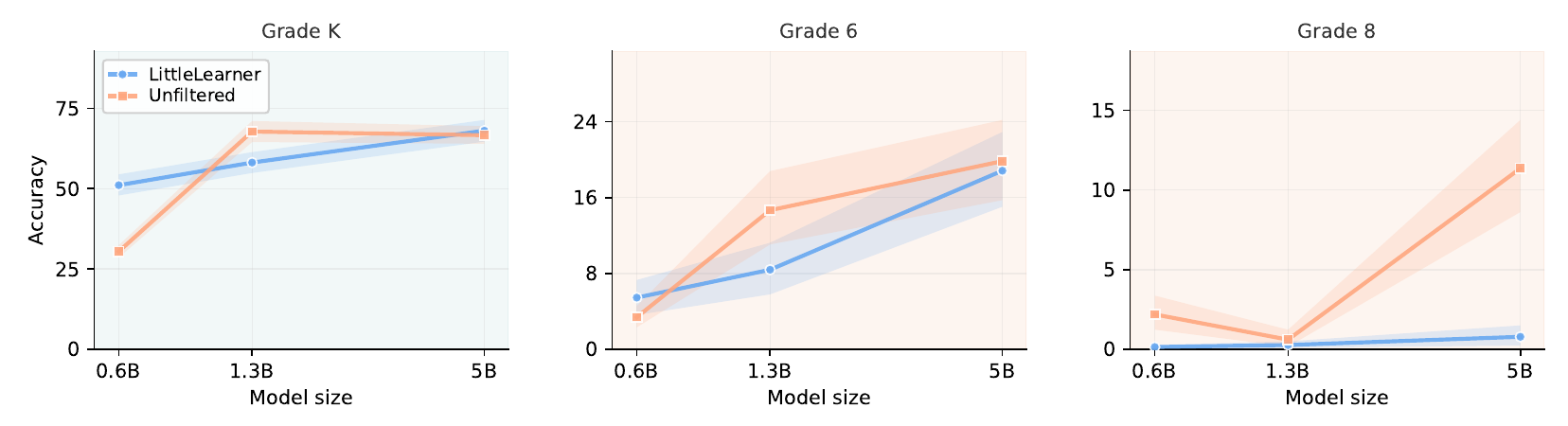}
    \caption{\textbf{Scaling model parameters helps performance within \KFive{}, and at the boundary, but does not recover performance well in \BKFive{}}. Performance of three differently sized \LittleLearner{}s and \General{} models on MathCAMPS. Shaded regions correspond to 95\% CIs. See \Cref{app:model-scaling} for all Grades.}
    \label{fig:model-size-comparison}
\end{figure*}

\subsection{Can Model Size Improve Out-of-Scope Performance?}

Model scale improves reasoning within the training exposure but does not meaningfully extend the model's capability boundary. To test whether additional capacity elicits skills outside the training distribution, we train \LittleLearner{} on \LittleCurriculum{} at 0.6B, 1.3B, and 5B parameters and evaluate each on MathCAMPS (\Cref{fig:model-size-comparison}).
Inside the exposure range scaling delivers substantial gains, and at the smallest size \LittleLearner{} actually outperforms the \General{} control: with no high school algebra or calculus competing for its limited capacity, the small model produces more focused reasoning traces. This specialization advantage disappears as the two models converge within \KFive{} at larger sizes, but their behavior diverges sharply in \BKFive{}. Scaling has a partial effect at the boundary of the training exposure (Grades 6 and 7), whose problem structure still overlaps with \KFive{} arithmetic, but essentially no effect on Grade 8, whose material lies fully outside \LittleCurriculum{}. \LittleLearner{} remains at floor across all three sizes. Curriculum exposure thus sets \LittleLearner{}'s capability frontier, and scale operates only within it.

\takeaway{Scaling model size improves performance within the model’s controlled knowledge exposure and extends modestly to problems along the same learning trajectory, but yields little improvement on problems requiring more advanced capabilities outside the exposure. See \Cref{fig:model-size-comparison}.}

\subsection{Can Post-Training through GRPO Recover Out-of-Scope Capabilities?}
\label{sec:post-training}

\begin{figure*}[t]
    \centering
    \includegraphics[width=\linewidth]{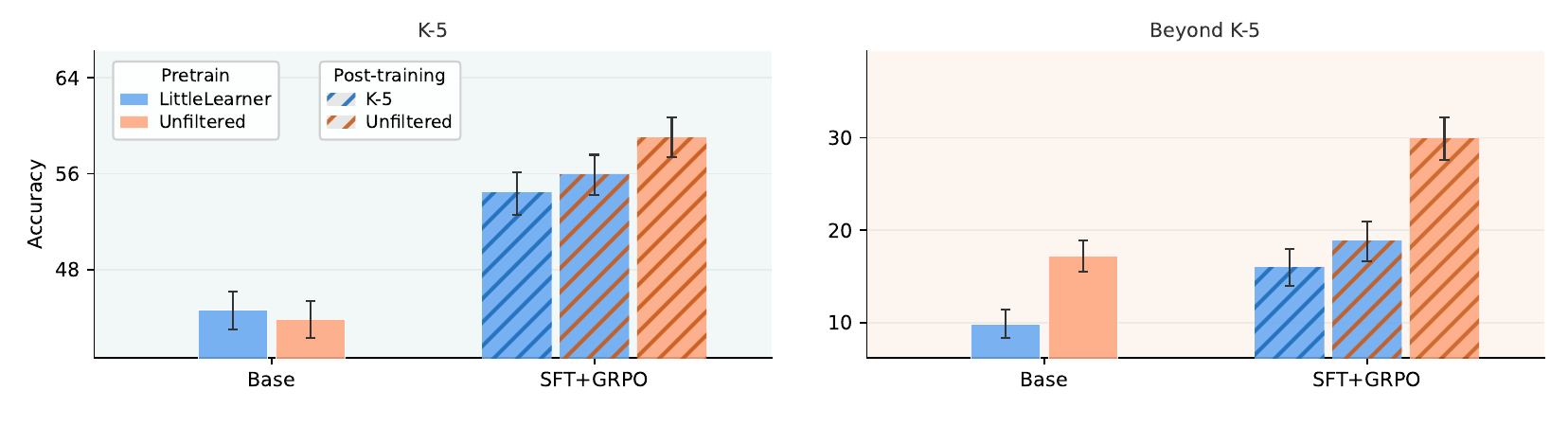}
    \caption{\textbf{Post-training on unfiltered content increases \KFive{} capabilities but does not overcome the \BKFive{} gap}. Performance on MathCAMPS for \LittleLearner{} and \General{} and their post-trained versions after SFT and GRPO. Whiskers represent 95\% CIs.
    }
    \label{fig:posttraining}
\end{figure*}

Both post-training on \KFive{} content as well as a matched unfiltered post-training baseline are unable to recover \BKFive{} performance for \LittleLearner{}. To isolate the effect of post-training with reinforcement learning (RL), we employ a two-stage pipeline with Supervised Fine-Tuning~(SFT) followed by GRPO~\citep{shao2024deepseekmath} (refer to \Cref{app:post-train-details} for details).
For \LittleLearner{} we first SFT on \KFive{} data selected using our filtering pipeline from \Cref{sec:data-filtering}; \General{} is SFT'd on a compute-matched unfiltered version of the same data sources.
At the GRPO stage we then ablate \LittleLearner{} between questions that passed our filtering, i.e., are from \KFive{}, and a compute-matched, unfiltered counterpart.
The same unfiltered dataset is used for GRPO on \General{}.
\Cref{fig:posttraining} shows MathCAMPS performance for all three runs.
Within \KFive{}, post-training lifts both models above their base performance with slightly larger gains for \General{}.
In \BKFive{}, post-training only modestly improves \LittleLearner{} while producing considerably larger gains for \General{}, widening the observed gap between the two models.
Crucially, within the tested post-training budgets, we observe no difference between \LittleLearner{} post-trained on \KFive{} data versus \BKFive{} data.
This highlights that bridging the performance gap between \LittleLearner{} and \General{} on out-of-scope content requires more fundamental changes beyond GRPO.

\takeaway{Post-training through GRPO significantly boosts in-scope \KFive{} capabilities, but fails to recover out-of-scope \BKFive{} capabilities, even when training with out-of-scope data. See \Cref{fig:posttraining}.}

\subsection{Can In-Context Learning Recover Performance on Out-of-Scope Tasks?}

We do not find that for the 5B \LittleLearner{} and the few-shot and explanation setups we test, in-context learning (ICL) is able to unlock new reasoning capabilities. Using the MathCAMPS setup, we synthesize three new problems per CCSS standard (\Cref{app:icl-question-generation}). We then manually author a chain-of-thought (CoT) solution for each question, which we use as few-shot examples (see \Cref{app:icl-generating-cot}).
Even though we observe that adding few-shot examples steers the model's output to be more aligned with the shots, e.g. through a shorter response (see \Cref{app:tab-different-icl}), this steering does not lead to better reasoning abilities. In \KFive{} using a few shots shows modest gains, and in \BKFive{} there is no gain at all. In \Cref{app:icl-diff-few-shots} we ablate different ways to present the few-shot examples and find that the natural hand-written examples perform the best.
When we add explanations instead of few shots, we do not observe the same behavior:
model responses do not change depending on the given explanation, and neither does performance.

\begin{figure*}[t]
    \centering
    \includegraphics[width=\linewidth]{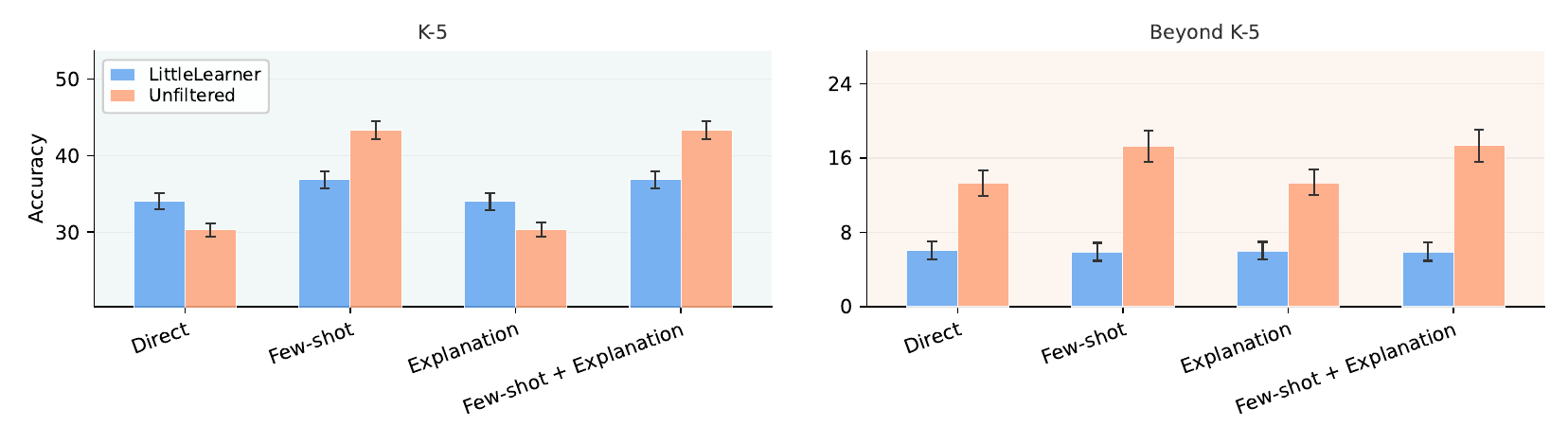}
    \caption{\textbf{In-context learning does not recover \BKFive{} capabilities}. \LittleLearner{} evaluated on MathCAMPS with few-shot evaluation (giving three example traces for each problem category) slightly improves \KFive{} performance, but not \BKFive{}. Adding explanations has no effect on performance. Whiskers indicate 95\% CIs. See \Cref{app:icl-diff-few-shots} for different few-shot examples.}
    \label{fig:icl}
\end{figure*}

\takeaway{In-context learning with the prompts we test does not unlock new reasoning capabilities in \BKFive{} for our trained 5B \LittleLearner{}. See \Cref{fig:icl}.}

\subsection{What Determines \LittleLearner's Capability Ceiling?}
Across all three interventions we study (model scaling, post-training with SFT+GRPO, and in-context learning), we observe a consistent pattern: each lever amplifies \LittleLearner's capabilities within its \KFive{} exposure, but provides limited gains in \BKFive\ on capabilities not well supported by pretraining. Scaling yields modest gains at the exposure boundary, but little improvement on more advanced out-of-scope material (\Cref{fig:model-size-comparison}); ICL does not unlock new reasoning capabilities \BKFive{} (\Cref{fig:icl}); and post-training amplifies in-scope capability for both bases but does not transfer to \BKFive{}, even when a \LittleLearner{} base is post-trained on the same unrestricted full-grade data that lift \General\ in \Cref{fig:posttraining}.
In each case, it is the pretraining filter, rather than the intervention, that sets the effective capability ceiling in our tested settings.

These findings are consistent with the line of work surveyed in \Cref{sec:rel}, which identified the \textit{pretraining distribution} as the dominant factor in downstream behavior. We view this not as a negative result but as a starting point: the same controlled exposure that exposes the limits of these standard interventions also makes \LittleLearner{} (and \LittleCurriculum) a useful sandbox for studying algorithms and learning paradigms that aim to expand model capabilities beyond their pretraining distribution. We sketch several such directions in the next section.

\section{Exploration Avenues: Beyond the \KFive{} Boundary}
We view the \LittleCurriculum{}--\LittleLearner{} sandbox as a playground to study how language models acquire and extend knowledge under explicit competency constraints.

\paragraph{RL, discovery, and extrapolation}
RL has become a central tool for extending language model capabilities beyond pretraining, yet in standard settings it is difficult to determine whether observed gains reflect newly acquired competencies or the re-elicitation of latent pretraining knowledge~\citep{yue2025doesreinforcementlearningreally,zhao2025echochamberrlposttraining}. Our setup offers a controlled alternative: because the model's prior is restricted to \KFive{} material, capabilities that emerge under RL, such as multi-digit arithmetic or multi-step algebraic reasoning, can be cleanly attributed to the RL process itself. We see two directions as particularly natural. First, \LittleLearner{} provides a small-scale analog of frontier discovery: studying whether reward signals, verifiers, self-play, and search can drive a bounded-knowledge agent toward higher-grade or competition-level mathematics offers a tractable proxy for the broader question of whether RL can produce genuine discovery in larger models. Second, the same setting supports controlled studies of algorithmic extrapolation, where RL pushes the model beyond its training range (for instance, from small-operand arithmetic to large numbers or longer sequences), providing a clean test of whether the model has acquired procedures rather than interpolated within its taught distribution.

\paragraph{Continual learning, interpretability, and calibration}
Continual learning, interpretability, and calibration are difficult to study cleanly in web-scale pretrained models, where broad prior exposure blurs the line between generalization, memorization, and latent knowledge. Our setup offers a controlled alternative: since \LittleLearner's exposure is explicitly specified, behavioral and representational changes can be related more directly to the concepts introduced during training. 
This suggests three immediate directions. First, introducing new material such as negative numbers or algebraic notation after the initial \KFive{} phase allows one to measure sample efficiency, retention, and interference between old and new knowledge. 
Second, the setup enables before-and-after interpretability studies, such as tracking how representations change when the model moves from whole numbers to fractions or from arithmetic expressions to symbolic equations. 
Third, tasks can be constructed within, near, or beyond the model's taught scope, enabling controlled evaluation of epistemic behavior: when the model answers correctly, expresses uncertainty, abstains, or hallucinates.

\paragraph{Educational science and human--model comparison}
Although \LittleLearner's boundary is defined through human curricula, it should not be interpreted as a model of a human child. Human and machine learners acquire language, concepts, and mathematical procedures through different mechanisms~\citep{mishra2024next,chang2022word}, and our results reflect this: \LittleLearner{} can sometimes perform better in downstream skills than in prerequisites  (see \Cref{app:more-details-mathcamps}). The developmental framing is therefore a tool for controlling exposure, not a claim of human-like development. This makes the sandbox useful as a complementary substrate for educational science: by specifying what the model has and has not seen, one can study how machines acquire skills under curriculum-like constraints and compare their trajectories to human learners along well-defined axes. For example, one could ask whether models and children require similar exposure to learn fractions, or whether they make similar errors on multi-step word problems. More broadly, \LittleLearner{} enables controlled human--model comparisons of sample efficiency, prerequisite dependence, transfer, and error structure.

\paragraph{Interactive and memory-based learning}
Evaluating these mechanisms in web-scale models is difficult because unknown training data obscures whether a capability is newly learned or previously memorized. \LittleLearner{} offers a controlled alternative: since its prior exposure is fully known, mechanisms can be assessed directly. We see three promising directions: First, testing retrieval and external memory to see if bounded models can reliably integrate out-of-domain content at inference. Second, using self-reflection and exploration to see how latent capabilities recombine without new data~\cite{kaur2026proper}. Third, studying multi-agent collaboration~\cite{kaur2026knowing} while precisely tracking the source of newly introduced information.

\section{Conclusion}
We introduce \LittleCurriculum{} and \LittleLearner{}, a developmentally constrained 88B-token corpus and 5B model that provide a controlled sandbox for studying knowledge acquisition.
In our experiments knowledge injection through post-training or in-context learning fails to unlock reasoning patterns or information from out-of-scope domains.
Although the 5B scale is small enough for academic feasibility yet large enough for coherent production, we acknowledge that certain emergent behaviors like in-context learning may be less pronounced than at frontier scales.
While our experiments instantiate this framework with \KFive{} exposure and a 5B learner, the same construction naturally extends to other pedagogically inspired constraints and model scales.  
Ultimately, \LittleCurriculum{} and \LittleLearner{} offer the academic community a tractable foundation for moving beyond the constraints of massive uncurated pretraining and toward controlled studies of true out-of-distribution knowledge acquisition.

% \newpage

\section*{Acknowledgements}

The authors would like to thank Timur Carstensen, Thomas Klein, Abhinav Menon, and Florian Windbacher for useful discussions.

This work was funded, in part, by the Collaborative Research Centre (CRC)
“Robust Vision – Inference Principles and Neural Mechanisms” of the German Research Foundation (DFG; SFB1233), project number 276693517. This work was additionally supported by the German Federal Ministry of Education and Research (BMBF): Tübingen AI Center, FKZ: 01IS18039A.
WB acknowledges financial support via an Emmy Noether
Grant funded by the German Research Foundation (DFG)
under grant no. BR 6382/1-1 and via the Open Philanthropy
Foundation funded by the Good Ventures Foundation. WB is a member of the Machine Learning Cluster of Excellence, EXC number 2064/1 – Project number 390727645. Authors acknowledge funding by the Federal Ministry of Research, Technology and Space of Germany (BMFTR, formerly BMBF) under grant no. 01IS24085C (OPENHAFM).
The authors thank the International Max
Planck Research School for Intelligent Systems (IMPRS-IS)
for supporting FL, TW, and PM. JZ and MPC are supported by the
Max Planck ETH Center for Learning Systems.

{
\small
\bibliographystyle{unsrtnat}
\bibliography{references}
}

% %%%%%%%%%%%%%%%%%%%%%%%%%%%%%%%%%%%%%%%%%%%%%%%%%%%%%%%%%%%%
\newpage
\appendix

\addcontentsline{toc}{chapter}{Appendix Contents}
\etocsetnexttocdepth{subsection}
\localtableofcontents

\section{Further Details on the Dataset Filtering Pipeline}

\subsection{Rule-Based Pre-Filtering}
\label{app:rule-based-pre-filtering}

\begin{figure}[h]
    \centering
    \includegraphics[width=\linewidth]{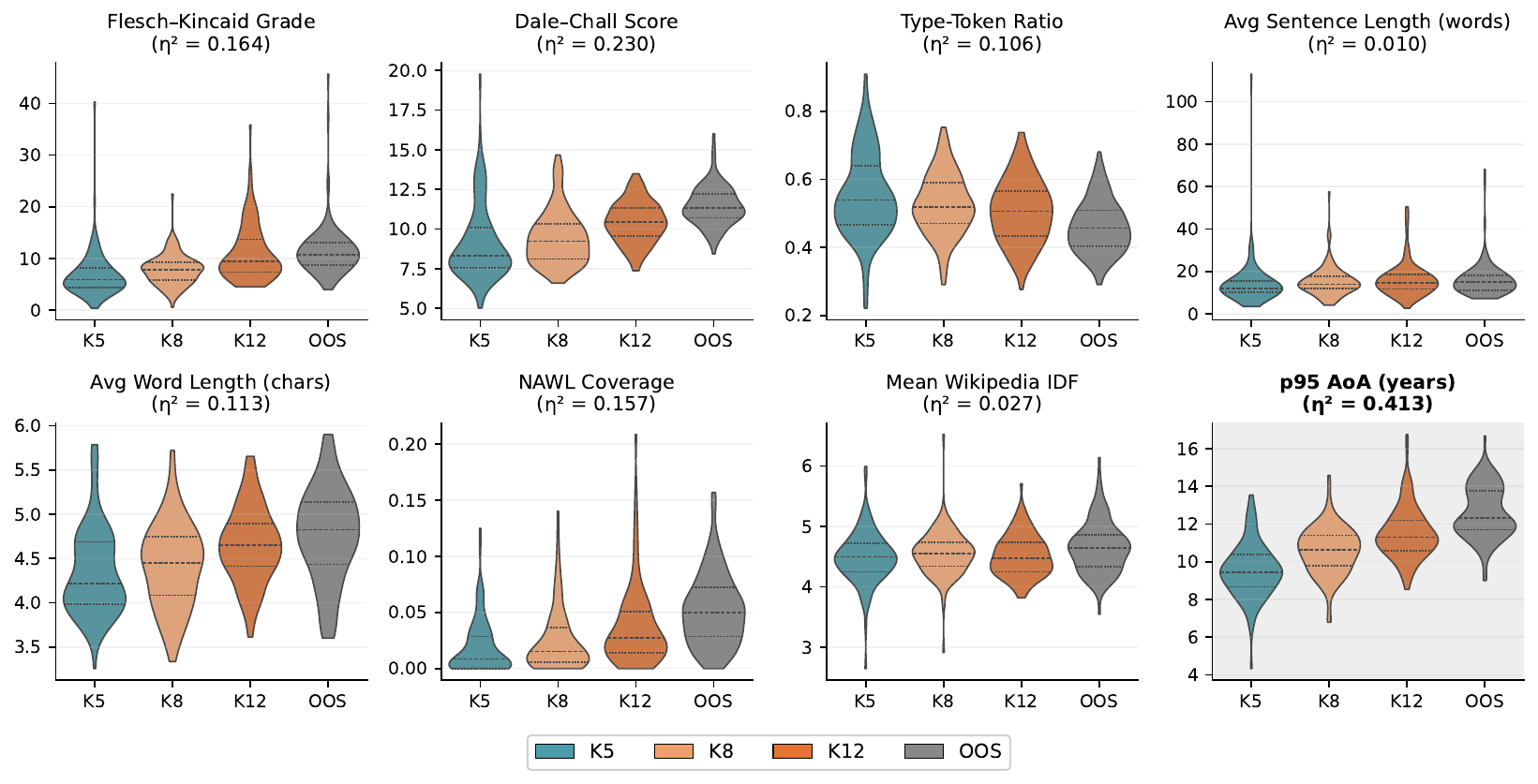}
    \caption{\textbf{Age-of-Acquisition gives the clearest signal for underlying Grade band}. Evaluating different rule-based metrics on our \CommonCoreText\ reveals that AoA is the best fit. K5 refers to K--5, K8 to middle school, K12 to high school and OOS to out-of-scope, i.e. beyond school level.}
    \label{fig:app-filtering-rule-based}
\end{figure}

\paragraph{Comparison of Rule-Based Metrics on \CommonCoreText} 
We evaluate different rule-based metrics to pre-sort \FinewebEdu{} (see \Cref{fig:app-filtering-rule-based}) and observe that among all tested rule-based approaches AoA gives the highest signal.
We test different readability metrics (Flesch--Kincaid~\citep{flesch1948new}, Dale--Chall~\citep{dale1948formula}), as well as simple linguistic metrics (Type-Token Ratio, Average Sentence / Word Length) and frequency metrics through the New Academic Word List (NAWL)~\citep{browne2013newawl} and IDF scores from Wikipedia~\citep{smartdataanalytics2020wikipedia_tfidf_dataset}.
Throughout all tested rule-based metrics, we observe that AoA leads to the best explanation of different grade bands.
However, it still does not explain the separation nearly well enough, and therefore we utilize a multi-stage approach.

\begin{figure}
    \centering
    \includegraphics[width=\linewidth]{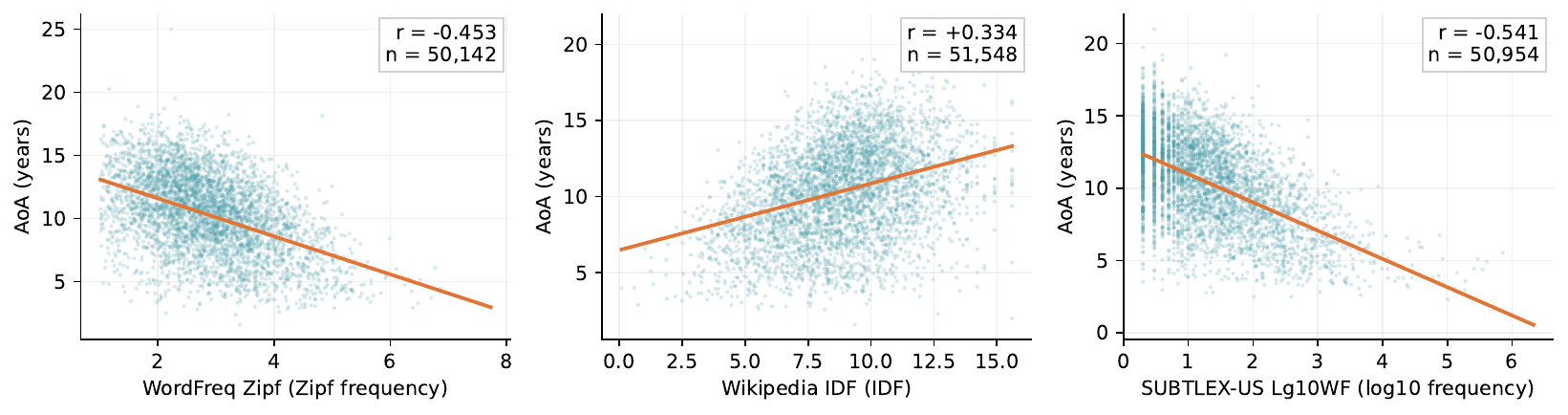}
    \caption{\textbf{At higher word coverage Word Freq Zipf gives the best trade-off between more word coverage and alignment to AoA}.}
    \label{fig:aoa_freq_corr}
\end{figure}

\paragraph{Filling In AoA Scores with Word Frequencies} 
Since \citet{kuperman2012age} report a high correlation between AoA scores and word frequency metrics, we check which word frequency metric is most aligned with AoA and also adds enough additional words to reduce the out-of-vocabulary (OOV) words from 10\% from AoA alone.
Although the word frequencies of SUBTLEX-US~\citep{brysbaert2009subtlexus} are the most aligned with AoA scores, 2.5\% of all words remain OOV.
Since this corpus consists of spoken text, the most frequent OOV words correspond to common written or web-specific artifacts, like www, uk, or facebook.
The word frequencies from Wikipedia cover most words that occur in \FinewebEdu{} (leaving only 0.23\% OOV). However, the alignment with AoA is low.
Therefore, we substitute missing AoA scores with WordFreq Zipf, as the tradeoff between alignment with AoA and new word coverage is the best (see \Cref{fig:aoa_freq_corr}).

\paragraph{OOV Words for AoA + Zipf} 
The most frequent words that do not appear in Zipf and AoA in (a subset of) \FinewebEdu{} are reported in \Cref{tab:app-oov-words-Zipf+AoA}.

\begin{table}[h]
    \centering
    \begin{tabular}{c|c}
        Token & Frequency (Percentage of Tokens) \\
        \hline
        chatgpt & 0.0031 \\
        mohill & 0.0026 \\
        doacs & 0.0017 \\
        sgas & 0.0016 \\
        phenprocoumon & 0.0016 \\
        shatavari & 0.0016 \\
        calfresh & 0.0014 \\
        necrozoospermia & 0.0014 \\
        srpf & 0.0013 \\
        fgas & 0.0012
    \end{tabular}
    \caption{\textbf{The most frequent out-of-vocabulary words for Zipf + AoA correspond to neologisms, misspellings or highly specialized terms}. We report the 10 most frequent out-of-vocabulary words for both AoA and Zipf on a subsample of \FinewebEdu.}
    \label{tab:app-oov-words-Zipf+AoA}
\end{table}

\subsection{LLMJ} 
\label{app:llmj_setup}

\begin{figure}
    \centering
    \includegraphics[width=\linewidth]{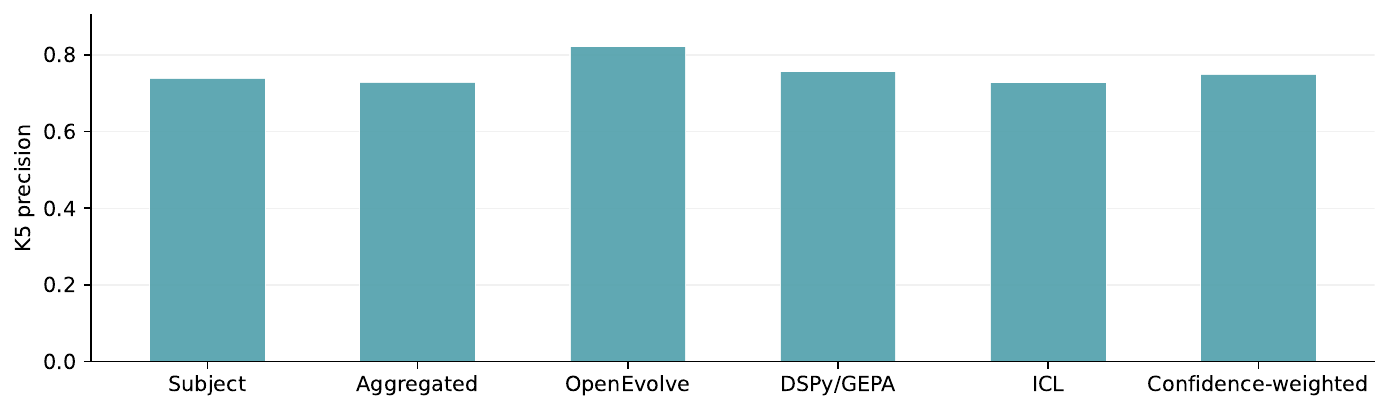}
    \caption{\textbf{No tested LLMJ method on Gemini 3 Flash vastly outperforms any other}. Since no tested LLMJ method vastly outperforms the others, we train classifiers on different labeled datasets and go with the setup yielding the best-performing classifier. Refer to \Cref{fig:llmj_ablation}.}
    \label{fig:llmj-setup}
\end{figure}

\paragraph{LLMJ Setup}
Our LLMJ pipeline classifies a text snippet into one of four grade bands (elementary school, middle school, high school, out-of-scope) by majority voting over multiple prompts queried from Gemini 3 Flash.
We instantiate three seed prompts targeting orthogonal signals: language complexity, context complexity, and a subject-aggregated prompt covering math, life science, physical science, and earth/space science.
All three are then optimized independently with OpenEvolve~\citep{openevolve} for 50 iterations using Gemini 3 Flash as both the task and evolution model, with a precision-on-validation fitness signal.
As an alternative regime, we also optimize a single chain-of-thought aggregated prompt with DSPy's GEPA optimizer~\citep{khattab2024dspy} (\texttt{auto=medium}, GPT-5 as reflection model, seed 42).
All judge calls use temperature~0 and a 1024-token cap.
Voting takes the majority label across the three setups (seed, OpenEvolve, DSPy); ties are broken toward the higher grade, since under-classification (e.g.\ surfacing high school content to a \KFive{} \LittleLearner) is more harmful than over-classification.

We additionally evaluated (i) in-context learning, adding two labeled examples per grade band to each prompt, and (ii) confidence-weighted voting, where each vote is weighted by the judge's self-reported confidence (\Cref{fig:llmj-setup}).
ICL matches the no-ICL setup in both accuracy and \KFive{} precision at roughly $2\times$ the cost; the reported confidences were poorly calibrated and confidence-weighting did not improve over uniform majority vote.
We also compared per-subject prompts against the aggregated prompt and found no significant accuracy difference, so we kept the aggregated form.

The per-sample inference cost is \$0.0014 for the OpenEvolve ensemble and \$0.0017 for the single GEPA-optimized prompt, versus \$0.0035 for the unoptimized seed prompts. This yields a total cost of \$0.0066 per sample for the entire ensemble or \$45,870,000 for all 6.95B samples in \FinewebEdu.

\begin{figure}
    \centering
    \includegraphics[width=0.8\linewidth]{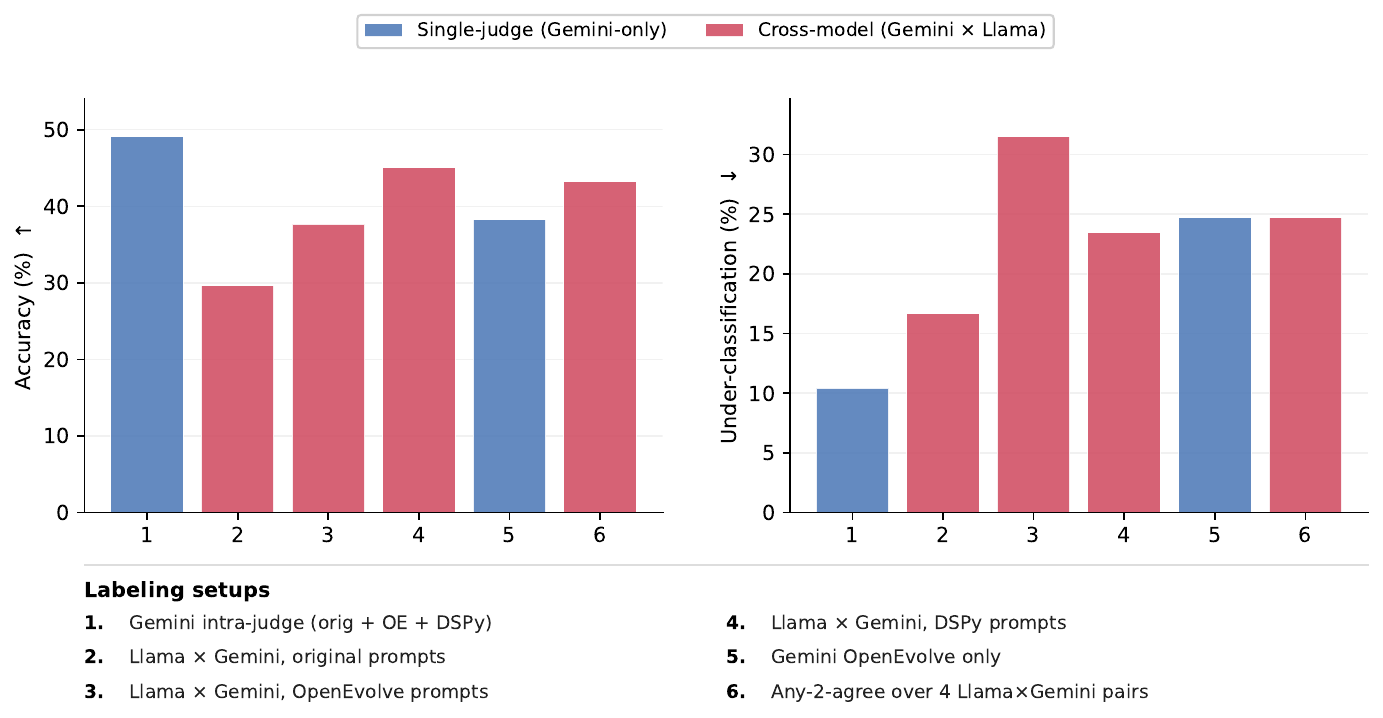}
    \caption{\textbf{The single-judge Gemini setup (1) outperforms cross-model agreement on both accuracy ($\uparrow$) and under-classification ($\downarrow$)}.}
    \label{fig:llmj_ablation}
\end{figure}

\paragraph{LLMJ Ablations with Training of Classifiers}
We compare single-judge labeling, specifically the Gemini-only setup with agreement across prompt variants, to cross-model aggregation based on Gemini–Llama agreement under multiple prompt variants. Single-judge setup (1) achieves the highest accuracy and lowest under-classification among all labeling setups in \Cref{fig:llmj_ablation}. We hypothesize that cross-model agreement removes ambiguous but informative examples near the decision boundary.

\begin{figure}
    \centering
    \includegraphics[width=0.8\linewidth]{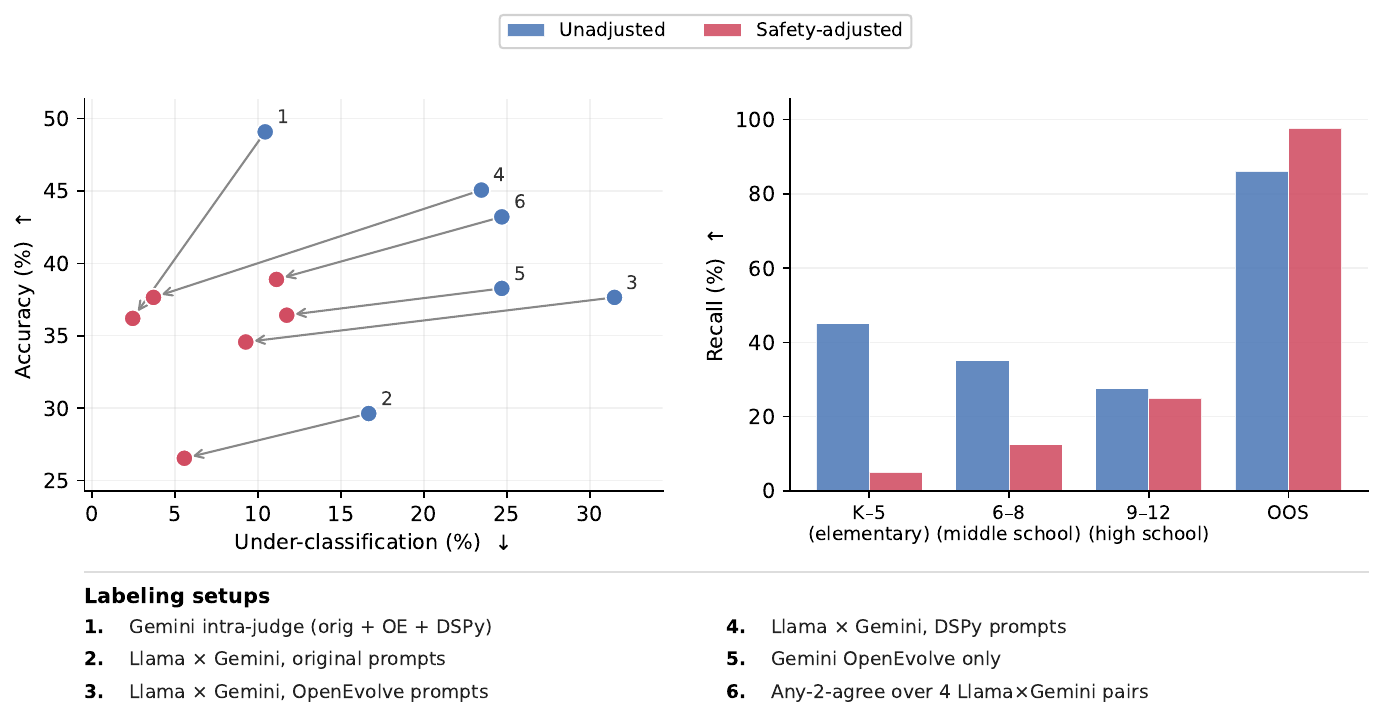}
    \caption{\textbf{The safety adjustment trades under-classification gain for a large accuracy loss and collapses recall on \KFive{} and 6--8 content}.}
    \label{fig:llmj_safety}
\end{figure}

\paragraph{LLMJ Safety Filtering} 
We also evaluate a safety adjustment that promotes predictions to higher grades when the cumulative probability mass above a threshold exceeds class-specific values.
Although this reduces under-classification, it does not improve overall reliability and degrades class-specific recall (\Cref{fig:llmj_safety}). We therefore retain the unadjusted predictions in the main results.

\subsection{Symbolic Filtering}
\label{app:symbolic-filtering}

Table~\ref{tab:symbolic-patterns} summarizes the notation families covered by the symbolic filter. Formal notation is sparse in the corpus, so the earlier lexical and semantic classifiers have limited evidence for learning these signals from LLMJ annotations. We therefore use high-precision rules for structured symbolic configurations, such as equations, exponents, roots, inequalities, functions, and calculus operators, rather than isolated symbols.

\begin{table}[t]
\centering
\caption{\textbf{Symbolic notation families targeted by the rule-based filter}. The table shows representative matches; each category also includes equivalent unicode, \LaTeX{}, and textual variants where applicable.}
\small
\begin{tabular}{p{0.26\linewidth}p{0.32\linewidth}p{0.34\linewidth}}
\toprule
Category & Examples & Pattern coverage \\
\midrule
Equations &
\(2x + 3 = 7\), \(ax + by = c\), \(x^2 + 2x\) &
Single-variable algebra, multi-variable equations, and polynomial forms \\

Exponents &
\(x^2\), \(x^n\), \(5^3\) &
Caret, \LaTeX{}, unicode, digit-base and programming-style \\

Roots &
\(\sqrt{x}\), \(\sqrt[n]{x}\), \(x^{1/3}\) &
Square roots, cube roots, \(n\)-th roots, \texttt{cbrt(}, and root word forms \\

Inequalities &
\(a < x < b\), \(a \leq x \leq b\) &
Chained comparisons with \(<\), \(>\), \(\leq\), and \(\geq\) \\

Calculus/operators &
\(\sum\), \(\int\), \(\partial\), \(\infty\) &
Summation, integrals, partial derivatives, infinity, and advanced \LaTeX{} commands \\

Functions &
\(f(x)\), \(g(y)\), \(h(z)\) &
Restricted to \(f/g/h\) with arguments \(x/y/z\) \\
\bottomrule
\end{tabular}

\label{tab:symbolic-patterns}
\end{table}

Across all symbolic patterns, the removal rate is approximately 0.1\% of the corpus after ModernBERT classification. The largest categories removed are polynomial expressions, exponent notation, and multi-variable equations, which are precisely the cases in which \BKFive{} content relies on formal notation rather than advanced vocabulary. The low removal rate indicates that this stage functions as a narrow cleanup pass rather than a general mathematics detector.
\subsection{Frequency Sampling}
\label{app:frequency-sampling}

Since the retained pool contains roughly 100B tokens, exceeding our 80B-token training target, we use the final sampling stage to tighten the exposure boundary rather than uniformly subsampling the retained documents. We use corpus frequency statistics to identify terms that are disproportionately associated with \BKFive{} material and preferentially exclude candidate documents that contain them.
For each term \(w\), we compute a contrastive frequency-ratio score between the retained \KFive{} pool and the \BKFive{} pool:
\[
\mathrm{score}(w)
=
\log_2
\frac{\mathrm{rate}_{\text{Beyond-K--5}}(w)}
{\mathrm{rate}_{\text{K--5}}(w)},
\qquad
\mathrm{rate}(w)
=
\frac{\mathrm{count}(w)}{\mathrm{total\ tokens}}.
\]

A positive score indicates that \(w\) is more characteristic of \BKFive{} material. We discard terms below a minimum frequency threshold to avoid unstable scores from rare strings. The remaining high-scoring terms are compiled into a rank-ordered blocklist, which is used to downsample candidate documents only in the final sampling stage.

\subsection{CommonCoreText}
\label{app:ccss-benchmark}

\paragraph{Construction}
\label{app:ccss-construction}
We evaluate the filtering pipeline on \CommonCoreText, a held-out benchmark of 460 passages sampled from textbooks and reading materials aligned with CCSS \citep{ccss2010}. Each passage is associated with a grade level and subject category from the source curriculum. The benchmark spans two domains: science (including mathematics, physical science, life science, and space science), and literature.\\

The corpus includes a mixture of curriculum-aligned materials (e.g., grade-level textbooks, science coursework, and standardized assessment content), as well as canonical literary texts drawn from CCSS exemplar and recommended reading lists (e.g., \textit{Alice's Adventures in Wonderland}, \textit{Charlotte's Web}, \textit{1984}, \textit{Macbeth}), ensuring coverage across both instructional and narrative text types.
This benchmark is held out from classifier training and is used exclusively to evaluate retention and specificity.

\paragraph{\KFive{} retained vs.\ removed analysis}
\label{app:ccss-k5-analysis}
Table~\ref{tab:k5_stats} reports summary statistics for retained and removed \KFive{} passages after filtering. Removed \KFive{} passages are consistently longer and more lexically complex than the retained passages, with the median length nearly doubling (115 to 205 words). This pattern aligns with the precision-oriented design of the pipeline, favoring simpler and clearly in-scope content.

\begin{table}[h]
\centering
\caption{Summary statistics for retained and removed \KFive{} passages after filtering.}
\label{tab:k5_stats}
\begin{tabular}{lcc}
\toprule
 & Retained & Removed  \\
\midrule
Word count (median) & 115 & 205 \\
Word count (mean) & 150 & 228 \\
Avg. sentence length (words) & 13.3 & 21.7 \\
Long words ($\geq$7 chars) & 12.3\% & 18.0\% \\
\bottomrule
\end{tabular}
\end{table}

\subsection{Independent Leakage Checks}
\label{app:leakage-audits}
\paragraph{Filtering on WeeBit}
To test whether our filter’s specificity holds outside CommonCoreText, we apply the full filtering pipeline without retuning to WeeBit \citep{vajjala-meurers}, a standard external readability corpus of graded educational texts consisting of Weekly Reader articles for younger grades and BBC Bitesize revision pages for older UK school stages (KS3/GCSE). WeeBit provides a clear gap around our boundary: Grades 2–4 (2,218 passages) and Grades 6–10 (6,000 passages). The retention rate on WeeBit provides a conservative upper bound on leakage: of the 6,000 \BKFive{} passages, our filter retains only 2.48\%, rejecting over 97\% of \BKFive{} text drawn from a corpus not used during filter development. Upon manual inspection, approximately half of these retained samples are web-scraping artifacts, and approximately a quarter consist of simple narrative prose. Of the remaining educational text, only three passages contain clearly out-of-scope concepts: fractions, exterior angles of polygons, and negative temperatures. These three passages correspond to 0.05\% of the full \BKFive{} split.

\paragraph{N-gram overlap of out-of-scope concepts}
To complement the WeeBit \citep{vajjala-meurers} evaluation, we conduct a corpus-level terminology audit. We compile 126 Grade-6+ technical n-grams drawn verbatim from the Common Core Standards for Mathematics and English Language Arts and the Next Generation Science Standards. The list is restricted to multi-word phrases (e.g., “kinetic energy,” “textual evidence,” and “Pythagorean theorem”) and excludes n-grams sharing vocabulary with the pipeline’s blocklist, keeping the audit separate from the filtering rules. Scanning the retained corpus, we find that only 0.09\% of passages contain at least one match.

Together, these checks provide complementary evidence that \BKFive{} leakage through the filtering pipeline is limited.

\subsection{Examples of Dropped Samples per Filtering Stage}
\label{app:exp_dropped_samples}

We report examples of discarded documents per filtering stage.
AoA Pre-filtering drops texts that are clearly outside the knowledge cut-off, discussing topics like neurology and cardiology.
The FastText and ModernBERT classifiers discard text with longer sentences, as well as discussing topics like the Gardner algorithm that are outside the knowledge cut-off.
The Symbolic Filtering specifically filters for mathematical symbols that are outside the knowledge cut-off like exponents, and roots.
Finally, frequency sampling makes sure that we sample from the middle of the remaining distribution.

\subsubsection*{Step 1: AoA Pre-filtering}
\begin{tcolorbox}[breakable]
\textbf{Example 1:}\\
{\ttfamily
In neuroscience, neuromodulation is the process in which several classes of neurotransmitters in the nervous system regulate diverse populations of neurons (one neuron uses different neurotransmitters to connect to several neurons). As opposed to direct synaptic transmission, in which one presynaptic neuron directly influences a postsynaptic partner (one neuron reaching one other neuron), neuromodulatory transmitters secreted by a small group of neurons diffuse through large areas of the nervous system, having an effect on multiple neurons. Examples of neuromodulators include dopamine, serotonin, acetylcholine, histamine and others.\\
A neuromodulator is a relatively new concept in the field, and it can be conceptualized as a neurotransmitter that is not reabsorbed by the pre-synaptic neuron or broken down into a metabolite. Such neuromodulators end up spending a significant amount of time in the CSF (cerebrospinal fluid), influencing (or modulating) the overall activity level of the brain. For this reason, some neurotransmitters are also considered as neuromodulators. Examples of neuromodulators in this category are serotonin and acetylcholine.\\

[...]
\medskip

Neuromodulation Other Uses\\
Neuromodulation also refers to a medical procedure used to alter nervous system function for relief of pain. It consists primarily of electrical stimulation, lesioning of specific regions of the nervous system, or infusion of substances into the cerebrospinal fluid. Electrical stimulation are devices such as Spinal Cord Stimulators (SCS) (surgically implanted) or transcutaneous electrical nerve stimulation devices (externally placed).\\}
\vspace{0.8em}

\textbf{Example 2:}\\
{\ttfamily 
Musings in the life of an internist, cardiologist and cardiac electrophysiologist.\\
Send them to the ER!\\
I think it is aortic aneurysm or mediastinal mass. It also shows hyperaeration which reflect chronic lung disease and possible smoking history which increase the risk of aneurysm or lung cancer. I also concern about right lower lobe patchy infiltrate, blunting of left costophrenic angle and haziness in right upper lung field close to mediastinum. I'm amazed to see "an asymptomatic patient" was hooked with telemetry and got portable CXR rather than regular PA and lateral which would yield much more information.More history and physical exam, previous CXR and obtaining CXR PA and lateral should help.\\
Is that the apex of the heart sticking up to the left? It looks like the cardiac shadow is sideways. I don't read xrays but is the diaphragm in the right place? Is "something" pushing the heart up and to the left?\\
Do a needle biopsy in the office.\\
It's not every day you see a descending aortic aneurism that exceeds 8 cm in diameter (confirmed by CT).Needless to say, we didn't do an atrial flutter ablation until the aneurism was addressed and no office-based needle biopsy was performed.:)\\
Post a Comment\\}
\end{tcolorbox}

\subsubsection*{Step 2: FastText Classification}

\begin{tcolorbox}[breakable]
\textbf{Example 1:}\\
{\ttfamily
First of all, I currently use a copper immersion chiller to cool my wort. It works, but takes at least 30 minutes to get below 100 and then i transfer and put it in my fermentation chamber till i reach pitching temp.\\
I am hoping to get a March Pump (like this but open to other recomendations http://goo.gl/Uo3pS) with the idea that i would pair this with a plate chiller in order to reduce the cooling time. I've never seen one in action, so my question is this: Do i need more equipment (other than hoses) beyond the plate chiller and pump to chill my wort? I already have a 1/2" ball valve off my keggle with a false bottom, so i expect all i need to do is get a hose from this valve down to the pump, out of the pump into the chiller, out of the chiller to the carboy. Am I missing anything? A Hopback? Any sage advice would be appreciated.
}
\vspace{0.8em}

\textbf{Example 2:}\\
{\ttfamily 
You have 12 coins, one of which is counterfeit, and a balance pan scale. The fake coin may be identified by the fact that it's weight
is different from the 11 genuine coins. Can you identify the
counterfeit coin and whether it is heavier or lighter in three
Background \& Techniques
This is an old problem with lots of literature available on the
web. The best reference is probably from one of Martin
Gardner's Mathematical Recreations columns published in
Scientific American magazine about 40 years ago. It apparently was
reprinted in book form in "Sixth Book of Mathematical
Games from Scientific American" which is
unfortunately out of print. I have copied a pretty good
summary of the algorithm from the alt.rec.puzzles newsgroup
to this page:\\
This program implements the Gardner algorithm to solve the problem (or check your solution) for this problem or the simpler case when you know whether the coin is heavy or light. .\\

[...]
\medskip

Notes for Programmers\\
A TScale object descended from TImage is used to define
the scale drawing methods and the animation depending on the relative
weights of the pans. An OnCountWeighing exit is used to
inform the parent program when a weighing should be counted. A
weighibg is ciunted whenever the number of weights in each pan is
moved the TScale definition stuff to a unit named U\_Scale just to
simplify browsing the main form unit, U\_Counterfeit.
U\_Counterfeit contains everything else required to let the user
select problem type and number of coins, and to drag the coins to
and from the balance pans and to the answer box. The CheckMinMoves
function provided the most challenge. It implements the Gardner
algorithms and builds the "cheat sheet" answer form, ResultsForm
defined in the U\_Results unit. This form is built
whenever the problem type or number of coins changes, but remains hidden
until the user click the "Show me button.
600+ lines of code in U\_Counterfeit and
300+ lines in U\_Scale are enough to put this well into the
Advanced category, but aside from CheckMinMoves, nothing is very
complex, just a lot of it.

}
\end{tcolorbox}

\subsubsection*{Step 3: ModernBERT Classification}

\begin{tcolorbox}[breakable]
\textbf{Example 1:}\\

{\ttfamily
When my son decided to stick a Lego up his nose earlier this summer requiring a trip to the emergency room, a lot of questions swirled through my mind: Is this normal? Why did he do this? Should I ban Legos from the house?\\
Then in early August a boy in Utah made headlines for having a Lego wheel removed from his nose. Apparently it had been there for years, causing breathing and sleeping problems.
All this activity got me wondering — what's the deal with kids jamming stuff up their noses?\\
"This is how children investigate their environment," said Dr. Jonathan Powell, a pediatrician with Resurrection Medical Group in Chicago. "When they are babies, they stick everything in their mouth. As they get a little older, they try other places. It's very common."\\
Dr. Michael Pitt, the director of the Pediatric Convenient Care Clinic at Lurie's Children's Hospital of Chicago, said this happens most often in kids between the ages of one and six.
"I think there's a misconception that boys do this more than girls but that's not the case," Pitt said. "For items found in the ears, it's equal between girls and boys, but for the nose, it's 2 to 1 (ratio) girls."\\

[...]
\medskip

Don't pretend you're an expert.\\
"If something is lodged in the nose, the first thing you can do is close their mouth and blow hard in the other nostril," Powell said. "Make sure they are sitting up and leaning forward. If that doesn't work, however, don't try anything else and get professional help immediately."\\
"Assume it's an emergency if you don't know what the item is that's gotten stuck," Pitt said. "A vast majority of the time the first line medical team can take care of it. You don't want to be taking out the tweezers and try to do this yourself. You could put it further up there and really make things worse."
}
\vspace{0.8em}

\textbf{Example 2:}\\
{\ttfamily 
Interactive Tool:How Bad Are Your Urinary Symptoms From Benign Prostatic Hyperplasia (BPH)?\\
BPH can be a bother, but it is usually not a serious health problem. If you are concerned about your symptoms, talk to your doctor about your options. They may include:\\
- Small changes to your lifestyle and urination habits, plus regular checkups to make sure that the problem is not changing or getting worse.\\
- Medicines to help reduce your symptoms.\\
- Surgery to help correct the problem and reduce your symptoms.\\
The main thing that helps you decide if and how to treat your symptoms is how much the symptoms bother you, not what your score is. To learn more, see these topics:\\
- Reference Benign Prostatic Hyperplasia (BPH)\\
- Opens New Window Enlarged Prostate: Should I Take Medicine? Opens New Window\\
- Opens New Window Enlarged Prostate: Should I Have Surgery? Opens New Window\\
Source: Roehrborn CG, et al. (2003, updated 2006). Guideline on the management of benign prostatic hyperplasia (BPH). American Urological Association Education and Research, Inc. Available online: http://www.auanet.org/content/guidelines-and-quality-care/.
|By:||Reference Healthwise Staff||Last Revised: Reference November 24, 2010|
|Medical Review:||Reference E. Gregory Thompson, MD - Internal Medicine
Reference J. Curtis Nickel, MD, FRCSC - Urology
}
\end{tcolorbox}

\subsubsection*{Step 4: Symbolic Filter}

\begin{tcolorbox}[breakable]
\textbf{Example 1:}\\
{\ttfamily
Looking for Cramster? Cramster is now Chegg Homework Help. Learn More
WILL RATE LIFESAVER!!!
Two parallel plates are connected to a 150-V power supply. The gap between the plates is filled with air, and sparks will fly if the electic field exceeds the breakdown value of \detokenize{E= 3.0 x 10^6 V/m}. What is the minimum spacing of the airgap (in mm)?
Anonymous answered1 minute later
You need a Homework Help subscription to view this answer!
}
\vspace{0.8em}

\textbf{Example 2:}\\
{\ttfamily 
A speeder passes a parked car at a constant speed of 30 \detokenize{m/s}. The
police car starts from rest at the instant the speeder passes him
and accelerates with a uniform acceleration of 2 \detokenize{m/s^2}.\\
a) What length of time does it take the police officer to catch up
with the speeder?\\
b) What distance do they both go until they are side by side?\\
c) What is the speed of the police car at the instant it catches up
to the speeder?\\
}
\end{tcolorbox}

\subsubsection*{Step 5: Frequency Sampling}

\begin{tcolorbox}[breakable]
\textbf{Example 1:}\\
{\ttfamily
The American Humanist Association advocates progressive values and equality for humanists, atheists and freethinkers in the United States. We work to promote humanism--the idea that you can be good without a god.\\
Darwin Day is a global celebration of science and reason held on or around Feb. 12, the birthday anniversary of evolutionary biologist Charles Darwin.\\
On this website you can find all sorts of information about Charles Darwin and the International Darwin Day Foundation. If you are hosting a Darwin Day event, you can post information about it on our events listing. You can also locate Darwin Day programs near you by searching our events section.\\
The AHA is proud to hold its 72nd Annual Conference in San Diego, CA, May 30-June 2, 2013 at the Bahia Resort Hotel. More details will be added soon. http://conference.americanhumanist.org/
}
\vspace{0.8em}

\textbf{Example 2:}\\
{\ttfamily 
Britain and Atlantis
Posted By: Charles Pope
Date: Sunday, 17 May 2009, at 10:06 a.m.
Excellent episode about stone age Britain today at 2 and 7 pm (EST) on History International Channel:
Britain was clearly a special place during the "Atlantean Age". This pre-history probably has much to do with the precedent of Britain being a stepping stone (ha, ha!) for greater kingship (ala Vespatian, Constantine, King John/Genghis Khan).
© Charles N. Pope, US Library of Congress. All rights reserved.
}
\end{tcolorbox}
\section{Details on Model Training}
\label{app:model-training}
\LittleLearner{} adopts the Qwen3-dense~\citep{yang2025qwen3technicalreport} architecture and is trained from scratch using Megatron-Core~0.17.1 with a custom training recipe~\citep{megatron-lm}. We train using 8-way layer-wise sharded Muon optimization~\citep{jordan2024muon} under a pure data-parallel configuration (TP=PP=CP=1), with BF16 parameter storage and MXFP8 computation. The cooloff mixture is composed of 91\% \KFive-filtered pretraining data, 5\% \KFive-rewritten mathematical reasoning data and 2\% \KFive-rewritten mathematical instruction-tuning data, both derived from Gemini~Flash~\citep{Gemini2_5FlashModelCard} rewrites of MegaMath-Pro~\citep{wang2025octothinker}, together with 2\% general instruction-tuning data rewritten by Gemini~Flash~\citep{Gemini2_5FlashModelCard} from held-out FineWeb documents not used during pretraining. Annealing positioning is used throughout the cooloff stage. All rewritten data are constrained to the \KFive{} curriculum to preserve the controlled knowledge boundary. 

To prevent leakage of \BKFive{} content through the tokenization pipeline, we train a custom tokenizer using only the \KFive-filtered \LittleCurriculum{} corpus. To test whether the \BKFive{} arithmetic gap is driven by number segmentation, we conduct a matched 2B ablation using single-digit number tokenization while holding all other settings fixed. \BKFive{} MathCAMPS pass@1 increases for both \LittleLearner{} (1.1\% $\rightarrow$ 1.8\%) and \General{} (4.6\% $\rightarrow$ 10.3\%), but \LittleLearner{} remains near floor and the qualitative gap persists.

We hold out a validation stream from the pretraining corpus, monitor validation bits-per-byte throughout training, and retain the checkpoint with the lowest validation bits-per-byte for downstream post-training and evaluation.

\section{Further Evaluation Details}

\subsection{Mathematical Reasoning with CoMTA}
\label{app:comta-binning}
The CoMTA dataset does not come with a CC-aligned label. Therefore, we label the dataset upon manual inspection, with Elementary corresponding to \KFive{}. Dialogs in this category discuss the multiplication of decimals (\texttt{5.NBT.B.7}), adding fractions (\texttt{5.NF.A.1}), and adding up to ten (\texttt{K.OA.2}).
Algebra is \BKFive{}, discussing matrix operations (\texttt{HSN-VM.A.2}) or working with negative exponents (\texttt{8.EE.A.1}). Trigonometry is also well above the K--5 level as trigonometric functions are discussed in high school (for example, in \texttt{HSF.TF.A.1}).
Calculus is also well outside the knowledge cut-off discussing, for instance, velocity and the derivative of logarithms.
We decided to exclude the Geometry category from our evaluation, since it contains both the calculation of simple volumes (\texttt{5.MD.C.5}) as well as trigonometric ratios, which is well above the K--5 knowledge cut-off.

\subsection{Mathematical Reasoning with MathCAMPS}
\label{app:more-details-mathcamps}

\subsubsection{Filtering MathCAMPS Standards and Questions}    
\label{app:mathcamps-filtering}
We remove a small subset of MathCAMPS standards and individual questions from our high-pass@$k$ evaluation, as these items do not reliably reflect a model's true reasoning capability. We apply three filters.  

First, we drop standards with fewer than 30 unique gold answers; these are predominantly multiple-choice items or questions whose answer space is small (perfect squares, cubes, and cube roots), and the resulting low entropy in the answer distribution inflates pass@$k$ regardless of the model's underlying reasoning capability.

Second, we discard individual questions where the gold answer string appears verbatim in the question text. Performance on these questions is up to $23\times$ higher than on otherwise comparable questions within the same standard; inspection of the model traces revealed that the model frequently restates the question without attempting to solve it, and the regex-based grader then registers the embedded gold answer as a correct response.

Finally, we completely exclude the CCSS standard $\texttt{6.EE.B.7}$.
The CCSS notation places this standard in grade 6 (``Solve real-world and mathematical problems by writing and solving equations of the form $x + p = q$ and $px = q$''), but the problems generated by MathCAMPS for this standard are structurally indistinguishable from grade-3 add/sub word problems under $\texttt{3.NBT.A.2}$.
For example, $\texttt{3.NBT.A.2-15-0}$ asks ``A restaurant started the week with 573 oranges \ldots they ended up using 228 oranges. How many oranges do they have left?'', while $\texttt{6.EE.B.7-18-0}$ asks ``A farmer had 414 grapes. After selling 78 of them, how many grapes do they have left?''. Including $\texttt{6.EE.B.7}$ would therefore double-count grade-3 add/sub competence in the grade-6 band, inflating any model's apparent \BKFive{} performance.

We then validate that the questions in \KFive{} and \BKFive{} have similar readability metrics, in order to rule out that a low performance on \BKFive{} stems purely from a linguistic shift rather than the conceptual shift we want to measure. \Cref{tab:mathcamps-readability} reports the scores for both partitions of MathCAMPS.

\begin{table}[ht]
\centering
\caption{Readability metrics for the \KFive{} and \BKFive{} partitions of MathCAMPS.}
\label{tab:mathcamps-readability}
\begin{tabular}{lcc}
\hline
\textbf{Metric} & \textbf{\KFive{}} & \textbf{\BKFive{}} \\
\hline
Flesch--Kincaid grade            & 5.28 & 5.37 \\
Mean sentence length (words)     & 10.7 & 10.1 \\
Mean word length (characters)    & 4.29 & 4.20 \\
Mean question length (words)     & 28   & 23   \\
\hline
\end{tabular}
\end{table}

\begin{figure}
    \centering
    \includegraphics[width=\linewidth]{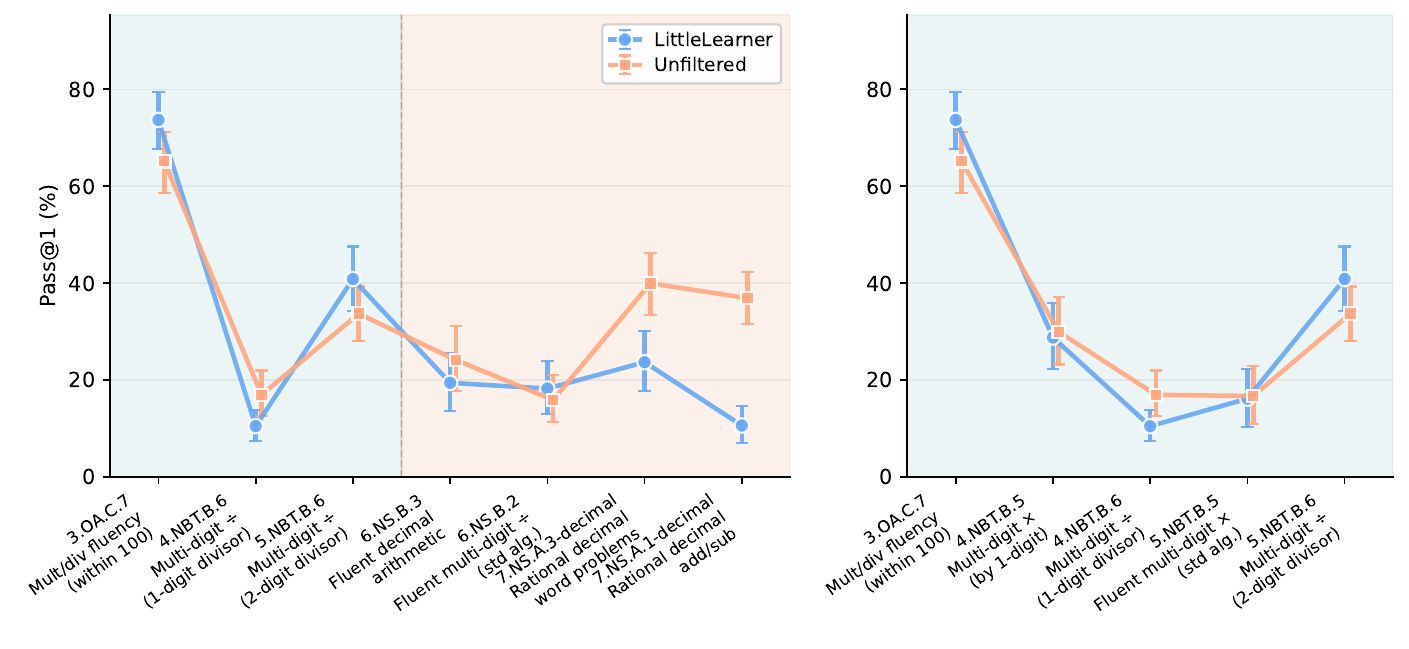}
    \caption{\textbf{\LittleLearner{} diverges more strongly on higher-grade arithmetic problems (left); \LittleLearner{} and \General{} do not follow CCSS learning paths (right)}. \LittleLearner{} and \General{} evaluated at pass@1 (estimated from 1024 samples) on MathCAMPS along learning paths in CCSS. Each standard on the left of the axis is a prerequisite for standards to the right on the axis. Left shows the division learning path; Right the division / multiplication learning path.}
    \label{fig:mathcamps-learning-paths}
\end{figure}

\subsubsection{Performance Across Learning Paths}
\label{app:mathcamps-learning-paths}
On the left panel of \Cref{fig:mathcamps-learning-paths}, \LittleLearner{} and \General{} track each other closely across Grades 3–6 and diverge in a framing-dependent way at Grade 7.
The similar performance on Grade 6 is not a failure of our filtering pipeline, but rather a mismatch between the human-aligned CCSS and the evaluation of models on these tasks. For instance, \texttt{6.NS.B.3} tests whether models solve arithmetic \emph{fluently}, which is hard to dissect for models. \texttt{6.NS.B.2} similarly tests whether students are able to solve division using the standard algorithm. This is not measured in MathCAMPS, as only final answers are evaluated.
\LittleLearner{} and \General{} diverge sharply only at Grade 7: \LittleLearner{}'s performance drops on rational decimal problems, specifically those framed as pure arithmetic (\texttt{7.NS.A.1-decimal}) compared to the corresponding word problems (\texttt{7.NS.A.3-decimal}), while \General{} is essentially framing-insensitive. This highlights that more \KFive{}-aligned framing does help \LittleLearner{}'s performance, but is not able to close the gap to \General{}.

The progression along CCSS prerequisite chains is non-monotonic. \LittleLearner{} is fluent at small mult/div (\texttt{3.OA.C.7}), drops on 1-digit-divisor multi-digit division (\texttt{4.NBT.B.6}), and then recovers on 2-digit-divisor division (\texttt{5.NBT.B.6}). Multi-digit multiplication follows a similarly jagged trajectory: 4th-grade multi-digit multiplication is easier (\texttt{4.NBT.B.5}) than its 5th-grade extension (\texttt{5.NBT.B.5}). Consistent with previous work~\citep{mishra2024next, chang2022word}, \LittleLearner{} (and \General{}) acquire skills along orderings that do not match the human curriculum. Therefore, \LittleLearner{} should not be read as a digital 12-year-old, but as a model with skills that arise from a \KFive{} restricted dataset, limited both by generic model capabilities and by the \KFive-restricted training data.

\subsubsection{Pass@k Performance on MathCAMPS for \LittleLearner{} and \General{}}
\label{app:mathcamps-pass-k}

\begin{figure}
    \centering
    \includegraphics[width=\linewidth]{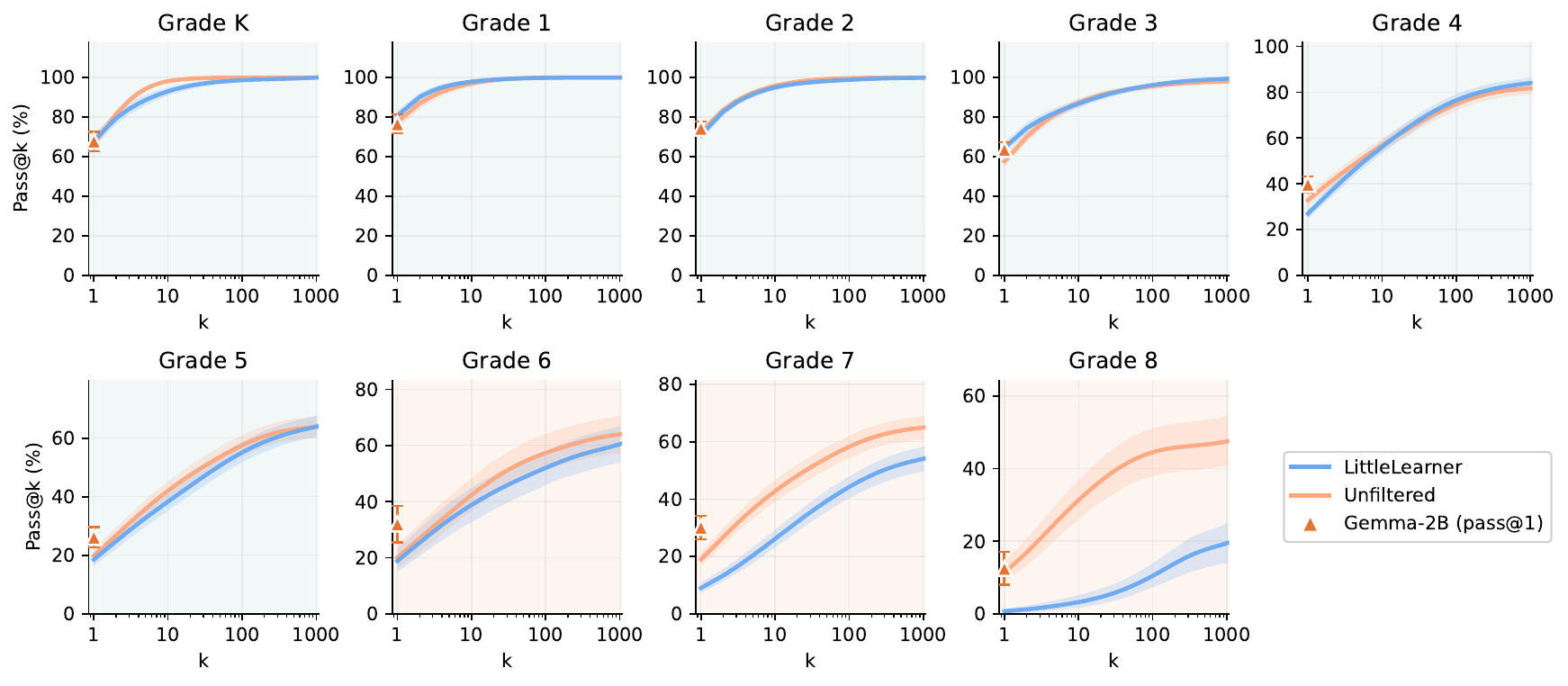}
    \caption{\textbf{The gap between \LittleLearner{} and \General{} on \BKFive{} remains stable for higher sampling budgets}. Performance of \LittleLearner{} and \General{} on MathCAMPS at higher pass@k. Shaded regions indicate 95\% CIs.}
    \label{fig:app-pass-at-k-curves}
\end{figure}

\LittleLearner{} is competitive with \General{} across \KFive{} but performance degrades more sharply as the evaluation moves into \BKFive{} (\Cref{fig:app-pass-at-k-curves}). At pass@1024, \LittleLearner{} retains essentially all of \General{}'s competence in Grades K--3, retains a comparable share through Grades 4--6, but only about two-thirds of it at Grade 8 with no overlapping CIs. The relative drop is consistent with the intended effect of \LittleCurriculum{}: filtering pretraining toward \KFive{} content preserves the elementary school capability while bounding extrapolation to later mathematical topics. This boundedness is visible not only in aggregate pass rates. Within \KFive{}, \LittleLearner{}'s per-problem successes are concentrated on the same problems \General{} solves reliably (Spearman = 0.80): the two models succeed and fail on largely the same items. In \BKFive{}, however, the pattern loosens: \General's per-sample success rate on problems \LittleLearner{} never solves rises to 5.1\%, roughly 4 times the equivalent \KFive{} rate, indicating \General{} can solve \BKFive{} questions that \LittleLearner{} cannot reach at any tested sample budget.

To probe whether \LittleLearner{}'s low \BKFive{} accuracy simply reflects an expression issue rather than a real capability ceiling, we evaluate under high sampling budgets with direct prompting, which yields high per-problem output diversity. On the log-scaled k-axis of \Cref{fig:app-pass-at-k-curves}, both models' curves flatten well before k=1024: gains slow noticeably by k$\approx$100 and are essentially zero from there to k=1024 in every panel. If \LittleLearner{} carried any latent \BKFive{} capability that a larger sample budget could surface, we are confident that it would have surfaced. The Grade-8 plateau is therefore best read as a capability ceiling of \LittleLearner{}'s training exposure, not as a sampling-budget artifact.

\subsection{Details on Scaling Experiments}
\begin{figure*}
    \centering
    \includegraphics[width=\linewidth]{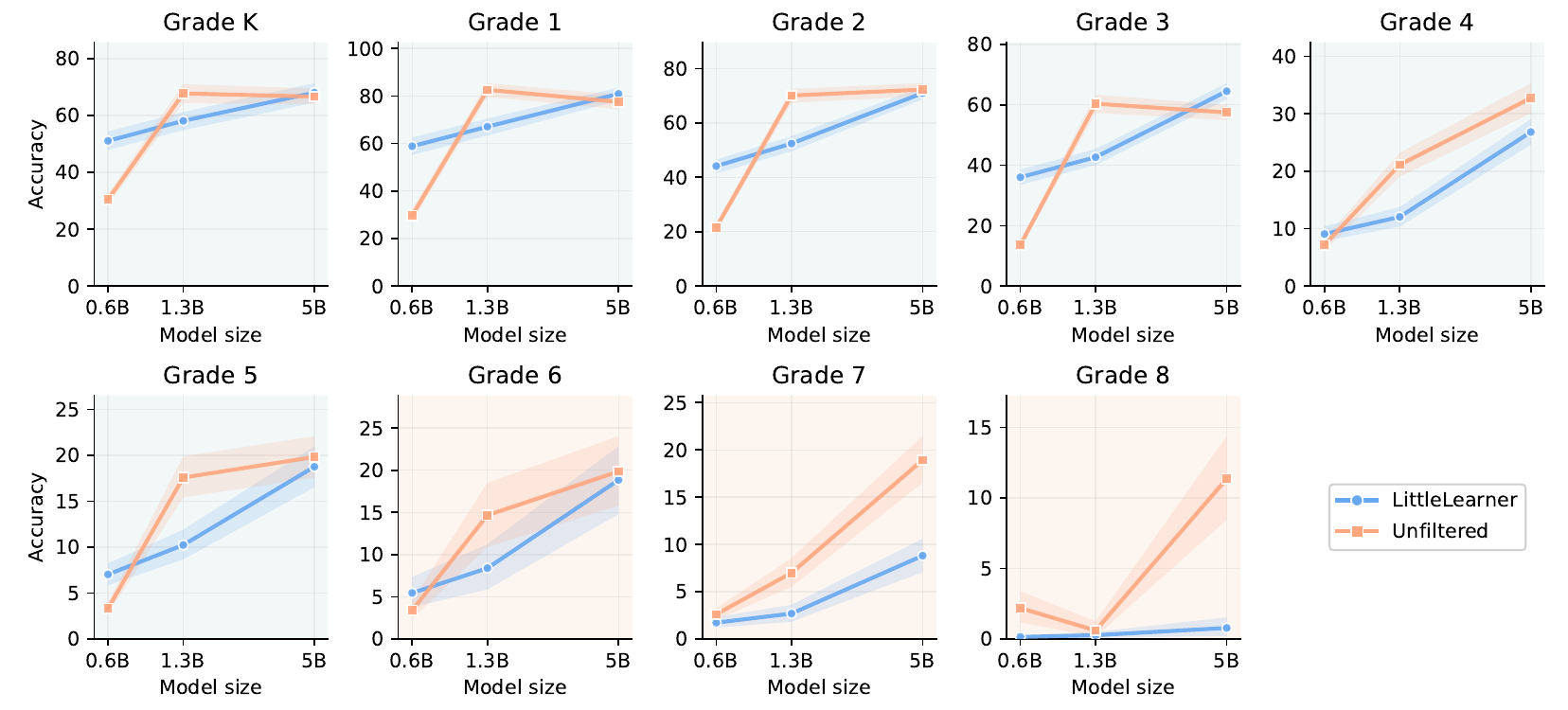}
    \caption{\textbf{Scaling model parameters helps performance within \KFive{}, and at the boundary, but does not recover performance well in \BKFive{}}. Performance of three differently sized \LittleLearner{}s and \General{} models on MathCAMPS. Shaded regions correspond to 95\% CIs.}
    \label{app:model-scaling}
\end{figure*}

\Cref{app:model-scaling} reports the scaling experiment numbers for each Grade. As outlined in \Cref{app:mathcamps-filtering}, the boundary in MathCAMPS between \KFive{} and \BKFive{} is fuzzy, reflected in scaling showing some effect in grade bands right at the boundary, whereas in Grade 8 (well within \BKFive{}) scaling shows no effect.

\subsection{Details on Post-Training Experiments}
\label{app:post-train-details}

The SFT stage uses the Gemini-rewritten chat data described in \Cref{app:model-training}: grade-school math word problems with chain-of-thought solutions (108k examples) and reading-comprehension question--answer pairs (526k examples). This results in ${\sim}46$M tokens used for a single epoch.
Note that pretraining runs fold a small fraction of this chat data into the final learning-rate decay phase (\Cref{app:model-training}); without it, neither base model is fine-tunable into a coherent chat model.

The GRPO stage is conducted on grade-stratified synthetic problems (generated via the MathCAMPS pipeline and checked for zero overlap with the evaluation set) alongside GSM8K~\citep{cobbe2021gsm8k}, in segments of 150--300 steps between which the training pool is re-banded to the problems the current policy solves 1--15 times out of 16. In the \KFive{} exposure of GRPO we limit GSM8K to problems that pass our filtering pipeline.

\subsection{Details on ICL Experiments}
\label{app:icl-variants}

\subsubsection{Question and Explanation Construction}
\label{app:icl-question-generation}
Questions are synthesized following MathCAMPS~\citep{mishra2024next}:
three questions per CCSS capability, generated with Gemini~3~Flash and
paired with a numerically verified answer. In parallel, we use the same
model with the Explanation Generation prompt (below) to produce a
grade-appropriate, procedural method-explanation per standard, using an
anchor block that names the standard, its grade, and which
\KFive{} prerequisites the student is assumed to know.

\prompt{Explanation Generation}{
Write a CONCRETE, ACTIONABLE method-explanation that teaches a step-by-step
procedure for solving any problem matching CCSS standard \{standard.id\}
(grade \{grade\}). Use only operations and vocabulary at or below grade 5.
The student's solutions look like numbered steps
(`Step 1: 6 - 1 = 5; Answer: 5`); your explanation should align with that
style.

Standard ID: \{standard.id\}
Grade level: \{grade\}
Full description: \{standard.description\}

\{anchor\_block\}

- 3 to 5 sentences (roughly 60-130 words). Be SPECIFIC and PROCEDURAL.
- State the method using CONCRETE NUMERICAL EXAMPLES. NEVER use letter
  variables OR generic placeholder letters (a, b, c, d, x, y, n, ...).
  GOOD: 'To multiply 3/5 x 2/7, multiply tops: 3 x 2 = 6; multiply
         bottoms: 5 x 7 = 35; the answer is 6/35.'
  BAD : 'To multiply a/b x c/d, the answer is (a x c) / (b x d).'
  BAD : 'Solve x + 5 = 12 by subtracting 5 from both sides.'
- Match how a grade-5 student writes work: numbered steps, plain ASCII
  (+ - x / =), fractions as a/b, mixed as 'N a/b'.
- If the procedure has cases, such as positive vs. negative inputs, state
  what to do in EACH case using a concrete example for each case.
- Connect the method to a K-5 mental model the student has, such as
  number-line jumps, inverse operations, parts of a whole, repeated addition,
  or area of a rectangle.
- Do NOT reference real-world scenarios, such as apples, dollars, or students.
- For unknowns, write '?' or 'the unknown number'; never a letter.
- BAN list — vague phrases that will be rejected:
   'as usual', 'in the usual way', 'look at whether', 'decide if',
   'depending on the type', 'the same way as before' without saying WHICH way,
   'just like normal multiplication' without naming the rule.
- Avoid grade-6+ vocabulary: 'rational', 'integer', 'variable',
  'coefficient', 'linear', 'system'. 'Equation' is OK; 'variable' is not.
- No LaTeX, no \textbackslash frac, no \$\$.

=== EXAMPLE OF A GOOD EXPLANATION ===
For 5.NF.B.4, multiplying fractions, shown for level and style only.
This example is DIFFERENT from the standard you are writing about.

Explanation: To multiply a fraction by a whole number, first rewrite the whole
number as a fraction over 1: 4 becomes 4/1. Then multiply the top numbers to
make the new top, and the bottom numbers to make the new bottom. For example,
4 x 3/8 = 4/1 x 3/8 = (4 x 3) / (1 x 8) = 12/8 = 1 4/8 = 1 1/2.
This works because multiplying by a whole number means repeating the fraction
that many times: 3/8 + 3/8 + 3/8 + 3/8 = 12/8.

Notice how the example uses ONLY concrete numbers (4, 3, 8) — no letter
placeholders, no variables — and connects the rule to repeated addition,
an idea the student already knows.

=== YOUR RESPONSE ===

Output exactly one line, prefixed with 'Explanation:' followed by your
4-to-6-sentence procedural explanation.
}

\prompt{Generated Explanation for \texttt{6.NS.B.2}}{
Method for this kind of problem:
To divide a large number like 3,432 by 12, first find how many
times 12 fits into the first two digits: 12 goes into 34 two
times because 12 $\times$ 2 = 24. Subtract 24 from 34 to get 10, then
bring down the next digit, 3, to make 103. Next, find how many
times 12 fits into 103: 12 $\times$ 8 = 96, so subtract 96 from 103 to
get 7. Finally, bring down the last digit, 2, to make 72, and
since 12 $\times$ 6 = 72, the final answer is 286. This method works
like an area model where you find the missing side of a
rectangle by taking away known chunks of the total area until
zero is left.
}

\subsubsection{Creating Hand-Authored Natural-Prose CoT}
\label{app:icl-generating-cot}
For the in-context demonstrations themselves, we hand-authored the
solutions rather than reusing model-generated CoTs. For each of the 43
\KFive{} CCSS standards used as anchors we wrote three worked examples
in a natural-prose style: a one-line orienting sentence
(``We need to find the answer.''), one sentence per computation using
connective language (``First,\ldots'', ``Then,\ldots''), and a
final \texttt{Answer:} line matching the format the eval harness parses.
The style deliberately matches what a
\KFive{} student explaining their own work would produce, and is close to
the free-form output the chat model produces without any exemplars. For instance:
\prompt{}{We need to find the answer.\\
    First, add the stones Leo found to the stones his sister gave him:
    $27 + 11 = 38$.\\
    Then, subtract the stones Leo gave to his friend from the total:
    $38 - 13 = 25$.\\
    Answer: 25}

\subsubsection{Comparing Different Few-Shot Examples}
\label{app:icl-diff-few-shots}

\paragraph{In-context example construction} We evaluate three families of in-context demonstrations that differ
\emph{only} in the CoT style; the question sets are identical. We outline the styles here with example traces for the question (2.NBT.B.5): \texttt{Leo found 27 smooth stones at the beach. His sister gave him 11 more
stones. Later that day, Leo gave 13 of his stones to his friend. How
many stones does Leo have now?}

\begin{itemize}
    \item \textbf{Compact algebraic} --- Terse, equation-only demonstrations. No natural-language framing, no wrapping sentences. Mean
    length: 99 characters. For instance: \prompt{}{3 tens + 7 tens = 10 tens\\
        $30 + 70 = 100$\\
        \#\#\# Answer: 100}

    \item \textbf{Natural prose} --- Hand-authored
    as noted above; mean length 374 characters, longest of the three. For instance: \prompt{}{We need to find the answer.\\
    First, add the stones Leo found to the stones his sister gave him:
    $27 + 11 = 38$.\\
    Then, subtract the stones Leo gave to his friend from the total:
    $38 - 13 = 25$.\\
    Answer: 25}

    \item \textbf{Q/A only} --- Ablation: same three exemplars
    with the CoT stripped, leaving only question and final answer.
    Mean length: 41 characters. For instance: \prompt{}{Answer: 100}
\end{itemize}

\paragraph{Performance Across Different ICL Strategies}
On the chat-tuned base~(\LittleLearner~5B~SFT), only natural-prose shots
improve over the zero-shot direct baseline; every other format strictly
hurts. See \Cref{app:tab-different-icl} for an overview of different performances.
Despite the model being able to follow the shots by adjusting output length, this does not result in better performance.

\begin{table}[t]
    \centering
    \caption{\textbf{None of the tested in-context learning shots can recover \BKFive{} performance, despite steering model behavior}. Performance of the SFT'd 5B \LittleLearner{} on MathCAMPS.}
    \label{app:tab-different-icl}
    \begin{tabular}{lccc}
    \toprule
    Method & \KFive{} & \BKFive{} & Mean output (chars) \\
    \midrule
    Direct (zero-shot) & $34.0\%$ & $6.0\%$ & 352 \\
    Natural prose & $36.8\%$ & $5.9\%$ & 374 \\
    Compact algebraic & $17.0\%$ & $1.6\%$ & 99 \\
    Q/A only & $9.2\%$ & $0.2\%$ & 41 \\
    \bottomrule
    \end{tabular}
\end{table}

\section{More Qualitative Examples}
\label{app:qualitative-base}

We compare generations from the \KFive-restricted \LittleLearner{} model and an unrestricted baseline model (\General) across prompts of increasing conceptual difficulty (\Cref{tab:app-qualitative-base}).

\begin{table}[h]
\centering
\caption{\textbf{Qualitative base-model examples highlight a clear distinction between in-scope and out-of-scope behavior}. Free-form completions from the released \LittleLearner{} base model and the matched \General\ base model on prompts spanning \KFive{} and \BKFive{} concepts.}
\label{tab:app-qualitative-base}

\begin{tabular}{p{0.23\textwidth} p{0.34\textwidth} p{0.34\textwidth}}
\toprule
\textbf{Prompt} & \textbf{\LittleLearner{}} & \textbf{\General} \\
\midrule

\multicolumn{3}{c}{\textbf{\KFive}} \\
\midrule

What is a magnet?
& A magnet is a piece of metal that is made to
attract or pull other metal pieces.
& A magnet is a material that has a property
called magnetism. \\[2pt]
\midrule

What is gravity?
& Gravity is the force that pulls everything down.
&
Gravity is the force between any two objects that is inversely proportional to the square of the separation between them... \\[2pt]
\midrule

\multicolumn{3}{c}{\textbf{\BKFive}} \\
\midrule

What is DNA?
& DNA is the instructions that tell us how we are going to look when we get old..
& DNA stands for deoxyribonucleic acid. It is a genetic molecule present in all living things, including humans. It is composed of two long strands of nucleotides… \\[2pt]
\midrule

What is Schrödinger's cat?
& Schrödinger's cat is a cat with two faces.
&
Schrödinger's cat is a thought experiment in quantum mechanics, proposed by Austrian physicist Erwin Schrödinger in 1935...\\[2pt]
\midrule

What does $E = mc^2$ mean?
& It's the name of a mathematical constant. It's also the name of a very famous character from the Star Wars movies.
& Einstein's famous equation says that energy is equivalent to mass times the square of the speed of light. \\

\bottomrule
\end{tabular}
\end{table}

In the in-scope prompts, \LittleLearner{} produces accurate explanations that reflect the simplified, progressively structured reasoning characteristic of a spiral curriculum, where concepts are introduced early and refined over time. Compared to the general model, its responses are often less detailed, but remain correct and developmentally appropriate.

In contrast, on out-of-scope prompts, the model does not express uncertainty but instead generates structured yet incorrect explanations. These responses are not random: rather than retrieving correct knowledge, the model systematically projects unfamiliar concepts onto familiar reasoning patterns learned during training.
For example, Schrödinger’s cat is reinterpreted as a literal cat with fabricated attributes.

This behavior suggests that the model lacks the underlying concepts, but nevertheless attempts to explain them using the available reasoning primitives, leading to coherent but incorrect outputs.

\newpage
\section{Existing Assets, Licenses, and Terms of Use}

Table~\ref{tab:asset_licenses} summarizes the existing assets used in this work,
their roles, license terms, and whether they are included in our release.
Where applicable, version information, access dates, and source URLs are provided
through the corresponding citations or release documentation.

\begin{table}[h]
\centering
\small
\setlength{\tabcolsep}{5pt}
\renewcommand{\arraystretch}{1.15}
\begin{tabular}{p{2.7cm} p{4.2cm} p{3.2cm} p{1.3cm}}
\toprule
Asset & Use in this paper & License / terms & Included? \\
\midrule
FineWeb-Edu & Source corpus for constructing \LittleCurriculum{}& ODC-BY & Yes \\
Common Core State Standards & Curriculum alignment & CCSS Public License & No \\
NGSS & Curriculum alignment & NGSS Public License & No \\
CLEAR & Readability validation & MIT License & No \\
CoMTA & Mathematical familiarity evaluation & Khan Academy Evaluation Dataset License & No \\
MathCAMPS & Mathematical reasoning evaluation & MIT License & No \\
Qwen3 & Training setup & Apache 2.0 & No \\
Gemma 2B & External baseline & Gemma Terms of Use & No \\
% Bespoke-Stratos & Post-training data & Apache-2.0 & No \\
% Reasoning Gym & Post-training data & Apache-2.0 & No \\
Smoltalk & Post-training data & Apache-2.0 & No \\
MMLU & Post-training data & Apache-2.0 & No \\
GSM8K & Post-training data & MIT & No \\
ARC & Post-training data & Apache-2.0 & No \\
\bottomrule
\end{tabular}
\caption{Existing assets used in the paper. “Included” indicates whether the asset itself, or a derived version of it, is distributed as part of our released artifacts. Unless otherwise indicated, third-party datasets, models, codebases, and curriculum standards are not redistributed.}
\label{tab:asset_licenses}
\end{table}

% %%%%%%%%%%%%%%%%%%%%%%%%%%%%%%%%%%%%%%%%%%%%%%%%%%%%%%%%%%%%

%newpage
% \clearpage
\section*{NeurIPS Paper Checklist}

\begin{enumerate}

\item {\bf Claims}
    \item[] Question: Do the main claims made in the abstract and introduction accurately reflect the paper's contributions and scope?
    \item[] Answer: \answerYes{} % Replace by \answerYes{}, \answerNo{}, or \answerNA{}.
    \item[] Justification: The main claims in the abstract and introduction accurately reflect the paper’s contributions and scope. The paper introduces \LittleCurriculum{} and \LittleLearner, and the experimental results support the central claim that controlled K--5 pretraining exposure creates an interpretable knowledge constraint that standard interventions such as scaling, post-training, and in-context learning do not substantially move beyond the training scope.

\item {\bf Limitations}
    \item[] Question: Does the paper discuss the limitations of the work performed by the authors?
    \item[] Answer: \answerYes{} % Replace by \answerYes{}, \answerNo{}, or \answerNA{}.
    \item[] Justification: The paper discusses limitations including model scale and the fact that \LittleLearner{} should not be interpreted as a digital human learner. These limitations help contextualize the empirical claims and clarify the scope of the conclusions.

\item {\bf Theory assumptions and proofs}
    \item[] Question: For each theoretical result, does the paper provide the full set of assumptions and a complete (and correct) proof?
    \item[] Answer: \answerNA{} % Replace by \answerYes{}, \answerNo{}, or \answerNA{}.
    \item[] Justification: The paper does not present theoretical results; it is an empirical dataset/model/constrained-pretraining study.

    \item {\bf Experimental result reproducibility}
    \item[] Question: Does the paper fully disclose all the information needed to reproduce the main experimental results of the paper to the extent that it affects the main claims and/or conclusions of the paper (regardless of whether the code and data are provided or not)?
    \item[] Answer: \answerYes{} % Replace by \answerYes{}, \answerNo{}, or \answerNA{}.
    \item[] Justification: The paper, appendix and supplementary material describe the filtering pipeline, validation data, model setup, evaluation datasets, post-training setup, and implementation details needed to reproduce or verify the main experimental results. The supplementary material includes the code and additional details required to reproduce the main experimental pipeline.

\item {\bf Open access to data and code}
    \item[] Question: Does the paper provide open access to the data and code, with sufficient instructions to faithfully reproduce the main experimental results, as described in supplemental material?
    \item[] Answer: \answerYes{} % Replace by \answerYes{}, \answerNo{}, or \answerNA{}.
    \item[] Justification:  The code is included in the supplementary material. The paper and appendix provide detailed descriptions of the data filtering pipeline, model training setup, post-training procedures, and evaluation protocols. While the final model checkpoints and processed datasets are being prepared for a permanent archival release (available upon publication), the provided code and documentation are sufficient to verify the experimental logic and methodology.

\item {\bf Experimental setting/details}
    \item[] Question: Does the paper specify all the training and test details (e.g., data splits, hyperparameters, how they were chosen, type of optimizer) necessary to understand the results?
    \item[] Answer: \answerYes{} % Replace by \answerYes{}, \answerNo{}, or \answerNA{}.
    \item[] Justification: The paper specifies the key experimental settings, including data filtering stages, classifier and LLM-judge setup, validation datasets, model training setup, MathCAMPS filtering rules, CoMTA labeling, post-training methods, and ICL variants. Additional data filtering pipeline, training and evaluation details are provided in the appendix.

\item {\bf Experiment statistical significance}
    \item[] Question: Does the paper report error bars suitably and correctly defined or other appropriate information about the statistical significance of the experiments?
    \item[] Answer: \answerYes{} % Replace by \answerYes{}, \answerNo{}, or \answerNA{}.
    \item[] Justification: The paper reports 95\% bootstrap confidence intervals for the post-training experiments, where uncertainty is most important for comparing training conditions. For other diagnostic figures, we report controlled empirical comparisons and provide the corresponding evaluation protocol and sample counts in the appendix.

\item {\bf Experiments compute resources}
    \item[] Question: For each experiment, does the paper provide sufficient information on the computer resources (type of compute workers, memory, time of execution) needed to reproduce the experiments?
    \item[] Answer:  \answerYes{} % Replace by \answerYes{}, \answerNo{}, or \answerNA{}.
    \item[] Justification: The paper reports the main compute resources for model pretraining, including the hardware and wall-clock time used to train \LittleLearner. The remaining experiments, including post-training, filtering, and inference-only evaluations, are substantially smaller in scale and are described at the level needed to understand the experimental setup.
    
\item {\bf Code of ethics}
    \item[] Question: Does the research conducted in the paper conform, in every respect, with the NeurIPS Code of Ethics \url{https://neurips.cc/public/EthicsGuidelines}?
    \item[] Answer: \answerYes{} % Replace by \answerYes{}, \answerNo{}, or \answerNA{}.
    \item[] Justification: The research conforms to the NeurIPS Code of Ethics. The work studies controlled language-model training and evaluation using documented datasets and experimental procedures, and we are not aware of any aspects of the research that require deviation from the Code of Ethics.

\item {\bf Broader impacts}
    \item[] Question: Does the paper discuss both potential positive societal impacts and negative societal impacts of the work performed?
    \item[] Answer: \answerYes{} % Replace by \answerYes{}, \answerNo{}, or \answerNA{}.
    \item[] Justification: The paper discusses the broader scientific value of the \LittleCurriculum–\LittleLearner sandbox for studying generalization, continual learning, interpretability, evaluation, and human–model comparisons under controlled knowledge exposure. The work may have positive impacts by enabling more rigorous studies of model knowledge restrictions and calibration.
    
\item {\bf Safeguards}
    \item[] Question: Does the paper describe safeguards that have been put in place for responsible release of data or models that have a high risk for misuse (e.g., pre-trained language models, image generators, or scraped datasets)?
    \item[] Answer: \answerYes{} % Replace by \answerYes{}, \answerNo{}, or \answerNA{}.
    \item[] Justification: The released model and dataset are designed for research on controlled knowledge exposure rather than deployment. The corpus construction pipeline intentionally filters content to a restricted K–5 scope, and the model’s limited scale reduce the risk of misuse compared with general-purpose pretrained large language models.

\item {\bf Licenses for existing assets}
    \item[] Question: Are the creators or original owners of assets (e.g., code, data, models), used in the paper, properly credited and are the license and terms of use explicitly mentioned and properly respected?
    \item[] Answer: \answerYes{} % Replace by \answerYes{}, \answerNo{}, or \answerNA{}.
    \item[] Justification: The paper credits the existing datasets, models, codebases, and curriculum standards used in the study through citations. \Cref{tab:asset_licenses} provides an asset table listing each asset’s role, version or access date, license or terms of use where available, and whether the asset is included in our release. 

\item {\bf New assets}
    \item[] Question: Are new assets introduced in the paper well documented and is the documentation provided alongside the assets?
    \item[] Answer: \answerYes{} % Replace by \answerYes{}, \answerNo{}, or \answerNA{}.
    \item[] Justification: The paper introduces new research assets, including \LittleCurriculum and \LittleLearner, and the associated construction pipeline. These assets are documented through descriptions of the corpus construction process, model training setup, evaluation protocols, intended research use, and known limitations.

\item {\bf Crowdsourcing and research with human subjects}
    \item[] Question: For crowdsourcing experiments and research with human subjects, does the paper include the full text of instructions given to participants and screenshots, if applicable, as well as details about compensation (if any)? 
    \item[] Answer: \answerNA{} % Replace by \answerYes{}, \answerNo{}, or \answerNA{}.
    \item[] Justification: The paper does not involve crowdsourcing experiments or new research with human subjects.

\item {\bf Institutional review board (IRB) approvals or equivalent for research with human subjects}
    \item[] Question: Does the paper describe potential risks incurred by study participants, whether such risks were disclosed to the subjects, and whether Institutional Review Board (IRB) approvals (or an equivalent approval/review based on the requirements of your country or institution) were obtained?
    \item[] Answer:\answerNA{} % Replace by \answerYes{}, \answerNo{}, or \answerNA{}.
    \item[] Justification: The paper does not involve human-subjects research or newly collected participant data.

\item {\bf Declaration of LLM usage}
    \item[] Question: Does the paper describe the usage of LLMs if it is an important, original, or non-standard component of the core methods in this research? Note that if the LLM is used only for writing, editing, or formatting purposes and does \emph{not} impact the core methodology, scientific rigor, or originality of the research, declaration is not required.
    %this research? 
    \item[] Answer: \answerYes{} % Replace by \answerYes{}, \answerNo{}, or \answerNA{}.
    \item[] Justification: The paper uses Gemini 3 Flash for grade-band annotation in the LLM-as-a-judge pipeline, prompt optimization, and curriculum-aligned labeling of evaluation questions, and these uses are documented in the method and appendix.

\end{enumerate}

\end{document}